\documentclass{article} 
\usepackage{iclr2027_conference,times}

\usepackage{amsmath,amsfonts,bm}

\def\eqref#1{equation~\ref{#1}}

\def\1{\bm{1}}

\DeclareMathAlphabet{\mathsfit}{\encodingdefault}{\sfdefault}{m}{sl}
\SetMathAlphabet{\mathsfit}{bold}{\encodingdefault}{\sfdefault}{bx}{n}

\usepackage{hyperref}
\usepackage{url}

\usepackage{graphicx}
\usepackage{booktabs}
\usepackage{amsmath,amssymb,amsthm}
\usepackage{algorithm}
\usepackage{algorithmic}
\usepackage{multirow}
\usepackage{xcolor}

\newcommand{\capsule}[4]{\mathcal{C}(#1, #2, #3, #4)}   
\newcommand{\dcos}{d_{\cos}}
\newcommand{\nlow}{n_{\mathrm{low}}}
\newcommand{\nhigh}{n_{\mathrm{high}}}
\newcommand{\normband}{N_b}
\newcommand{\conceptgeo}{\Gamma}
\newcommand{\CapsuleLens}{Capsule Lens}

\newtheorem{definition}{Definition}

\title{Capsule Lens: Locating and Tracking Concept Geometry in Model Representations}

\author{Yiming Tang\thanks{Equal contribution. Email: \texttt{yiming@nus.edu.sg}} \\
National University of Singapore \\
\And
Harshvardhan Saini\footnotemark[1]\\
Indian Institute of Technology (ISM) Dhanbad \\
\And
Samyak Jha \\
Indian Institute of Technology (ISM) Dhanbad \\
\And
Huaming Chen\\
The University of Sydney \\
\And
Xufeng Duan \\
The Chinese University of Hong Kong \\
\And
Dianbo Liu\thanks{Corresponding author. Email: \texttt{dianbo@nus.edu.sg}} \\
National University of Singapore
}

\iclrfinalcopy 
\begin{document}

\maketitle

\vspace{-4mm}
\begin{abstract}
Understanding how concepts are encoded in the internal representations of machine learning models is a central problem in mechanistic interpretability, essential both for the science of deep learning and for the trustworthy deployment of increasingly capable models. Existing approaches to interpret model representations mainly map representations onto more interpretable spaces and do not directly characterize how concepts occupy representation space; various hypotheses have been proposed, but often lack of rigorous validation and largely focus on static representations. In this work, we introduce \textbf{Capsule Lens}, a framework that matches the region a concept occupies with a simple, trackable geometric form, a \emph{capsule}, defined by several interpretable parameters, fitted in closed form to each concept's geometry and validated on held-out samples. We apply \CapsuleLens{} in two major settings: static and dynamic representations. On static representations, we demonstrate how to locate concept geometry across various models, and how the span and norm curves uncover important geometric characteristics. On dynamic representations, we present three case studies tracking representation drifts induced by distinct training settings, CLIP pretraining, RL post-training on visual question answering, and RL post-training on mathematical reasoning. These analyses reveal qualitatively different geometric dynamics, ranging from broad network-wide restructuring in CLIP pretraining to localized and concept-specific changes in RL post-training. Our results include findings aligned with existing literature as well as novel observations. We believe \CapsuleLens{} stands as a promising tool for locating, analyzing, and tracking concept geometry in both static and dynamic representations.
\end{abstract}

\vspace{-2mm}
\section{Introduction}
\label{sec:intro}

Understanding how concepts are encoded in the internal representations of neural networks is a central goal of mechanistic interpretability \citep{sharkey2025open}. As models achieve remarkable capabilities across diverse domains \citep{brown2020languagemodelsfewshotlearners, Guo_2025, grattafiori2024llama3herdmodels}, deciphering what they represent---and how---is important for both the science of deep learning and the trustworthy deployment of increasingly capable systems \citep{bereska2024mechanisticinterpretabilityaisafety}. Beyond this static picture, an equally important question is how these internal representations change during training. Post-training methods such as supervised fine-tuning and reinforcement learning are now deployed at scale \citep{ouyang2022traininglanguagemodelsfollow, shao2024deepseekmathpushinglimitsmathematical}, yet their effects on model internals remain poorly understood \citep{wang2025understandingfinetuningmechanismsllms,xu2025trackingfeaturedynamicsllm}. This gap is consequential, as improper fine-tuning can induce broad behavioral changes, including misalignment \citep{Betley_2026,dong2026rlpluscounteringcapabilityboundary}. Understanding such dynamics first requires locating concept geometry in representation space and then tracking how it changes during training.

To address these questions, researchers have developed interpretability lenses and neural geometry approaches, each addressing a different part of the problem. Interpretability lenses map internal activations onto more interpretable spaces: the Logit Lens and Tuned Lens map hidden states onto tokens \citep{nostalgebraist2020logitlens, belrose2025elicitinglatentpredictionstransformers}, Patchscopes and Rep2Text decode representations into natural language \citep{ghandeharioun2024patchscopesunifyingframeworkinspecting, zhao2025rep2text}, and sparse autoencoders decompose polysemantic activations into more interpretable features \citep{cunningham2023sparseautoencodershighlyinterpretable, gao2024scalingevaluatingsparseautoencoders, tang2025human}. These lenses reveal \emph{what} information is represented, but provide limited insight into \emph{how} a concept occupies representation space. Neural geometry instead studies the geometric structure of representations, proposing that concepts may be encoded as linear directions \citep{park2024linearrepresentationhypothesisgeometry, elhage2022toymodelssuperposition}, polytopes and hierarchies \citep{park2025geometrycategoricalhierarchicalconcepts}, or irreducibly multidimensional manifolds \citep{engels2025languagemodelfeaturesonedimensionally}. However, these accounts typically posit or probe geometric structures rather than directly estimating and validating the region occupied by a concept \citep{tang2026unifiedtheorysparsedictionary, bhalla2026sparseautoencoderscaptureconcept}. Moreover, both lines of work largely focus on static representations, providing limited insight into how concept geometry changes during training.

Following recent work \citep{xiong2026latticerepresentationhypothesislarge, modell2025originsrepresentationmanifoldslarge}, we formulate a concept as a binary indicator on a dataset: a sample either exhibits the concept or does not. Under this formulation, concept geometry is the region occupied by the concept's positive samples in representation space, approximated by a set of points \citep{sorscher2022neural, chung2021neural, bhalla2026sparseautoencoderscaptureconcept, wollschläger2026geometryrefusallargelanguage}. Our key observation is that understanding how a concept is encoded reduces to characterizing this geometry in $\mathbb{R}^d$, while understanding how training reshapes the encoding requires tracking how the geometry changes across checkpoints. Drawing inspiration from diverse geometric views of concept representations \citep{engels2025languagemodelfeaturesonedimensionally,modell2025originsrepresentationmanifoldslarge,sorscher2022neural}, we approximate each concept geometry with a simple parametric shape---a capsule---whose parameters can be directly compared across representations and model states.

In this work, we introduce \CapsuleLens{}, a framework that approximates the region occupied by a concept with a simple, trackable geometric form---a \emph{capsule}, defined by a unit axis, an axis span, and a norm band---fitted in closed form to each concept's representation cloud. We also introduce two coverage curves, the span curve and the norm curve, which reveal how representations populate the fitted region beyond its boundary parameters, together with a validation protocol that evaluates fitted capsules on held-out samples. We apply \CapsuleLens{} in two complementary settings. For static representations, we characterize concept geometry across various model architectures, examining its internal structure, disentanglement, and variation across model depth. For dynamic representations, we use three case studies to track how concept geometry changes under distinct training settings, spanning CLIP pretraining and RL post-training on visual question answering and mathematical reasoning. These studies reveal qualitatively different geometric dynamics: broad network-wide restructuring in CLIP pretraining and more localized, concept-specific changes in RL post-training. In particular, we observe a concept-specific concentration effect under RLVR, where the geometry of a small subset of concepts sharpens at particular layers while most concept regions remain nearly unchanged. We believe \CapsuleLens{} provides a practical tool for locating, characterizing, and tracking concept geometry across both static representations and training dynamics.

Our contributions are listed as follows:
\begin{itemize}
    \item We propose \CapsuleLens{}, a parametric, closed-form, and trackable lens on concept geometry, equipped with span and norm coverage curves.
    \item We locate concept geometry across diverse model architectures, characterizing its internal structure, disentanglement, and layer-wise evolution.
    \item Across three case studies, we uncover qualitatively different geometric dynamics across distinct training settings: broad network-wide restructuring in CLIP pretraining, and localized, concept-specific changes in RL post-training.
    \item We further identify a concept-specific concentration effect under RLVR, where a small subset of concept geometries sharpen at particular layers while most concepts remain nearly unchanged, revealing an underexplored form of representation change.
\end{itemize}

\vspace{-2mm}
\section{Method}
\label{sec:method}

In this section, we introduce our proposed framework, \CapsuleLens{}. Section~\ref{sec:preliminaries} introduces the formal definitions of concepts and capsules. Section~\ref{sec:matching} describes capsule matching, the closed-form procedure that fits a capsule to a concept's representation geometry, along with the span and norm curves that describe how the concept's representations distribute within the fitted region. Section~\ref{sec:validation-protocol} presents the held-out validation protocol used to evaluate fitted capsules. Section~\ref{sec:tracking} explains how we track concept geometry by comparing matched capsules across training checkpoints.

\begin{figure}[H]
    \centering
    \includegraphics[width=1\linewidth]{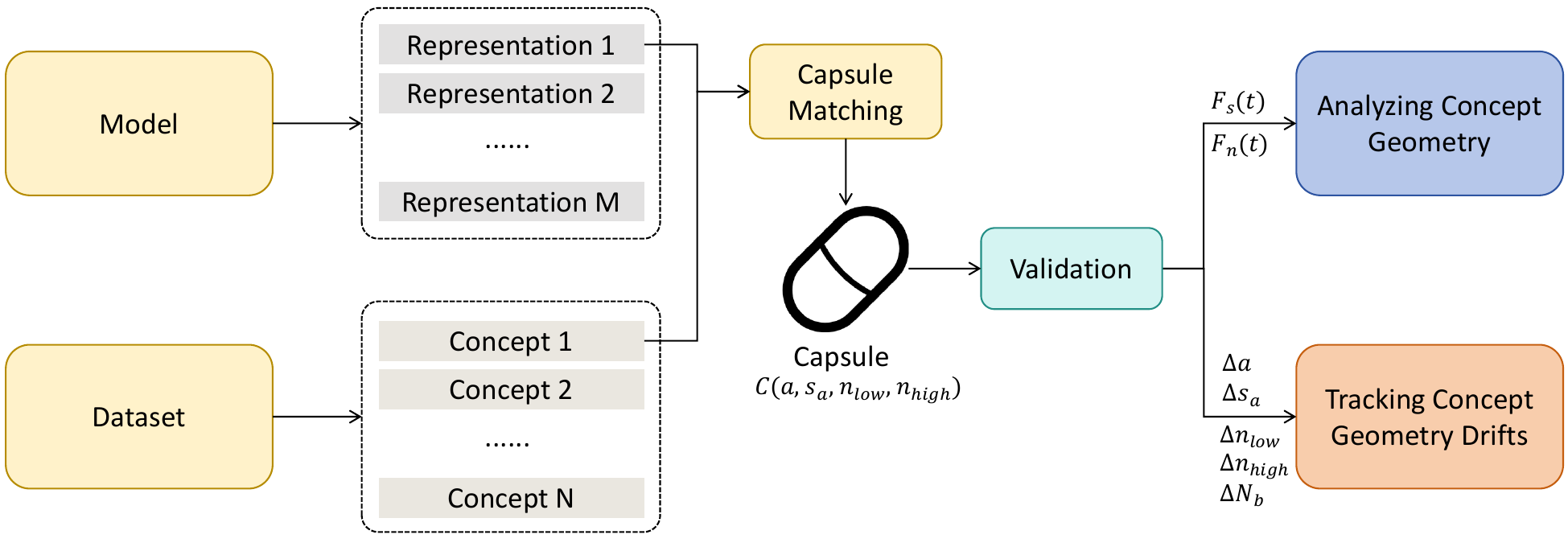}
    \vspace{-8mm}
    \caption{\textbf{Overview of the \CapsuleLens{} framework.} Given a model and a dataset, the framework extracts representations for a set of predefined concepts. Capsule matching fits a capsule \(\mathcal{C}(a, s_a, n_{\mathrm{low}}, n_{\mathrm{high}})\) to each concept's representation cloud in closed form. The fitted capsule parameters and the associated span curve \(F_s(t)\) and norm curve \(F_n(t)\) are used to analyze the static geometry of concepts. The drift metrics---axis rotation \(\Delta a\), span change \(\Delta s_a\), and norm-band changes \(\Delta n_{\mathrm{low}}, \Delta n_{\mathrm{high}}, \Delta N_b\)---are used to track how this geometry shifts across training checkpoints.}
    \label{fig:method}
\end{figure}

\subsection{Preliminaries: Concepts and Capsules}
\label{sec:preliminaries}

We work in the representation space $\mathbb{R}^d$. A model $\phi_\theta : \mathcal{D} \to \mathbb{R}^d$ maps each sample to a representation, e.g., a token's activations at a given layer of a large language model, or the visual embedding of an image encoder. For a nonzero $x \in \mathbb{R}^d$ and a unit vector $a \in \mathbb{R}^d$ ($\|a\| = 1$), the \emph{cosine distance} between $x$ and $a$ is
$\dcos(x, a) = 1 - \frac{\langle x, a \rangle}{\|x\|} \in [0, 2].$

\begin{definition}[Concept Geometry]
\label{def:concept}
A concept $c$ is a binary indicator $I_c : \mathcal{D} \to \{0, 1\}$ on a dataset $\mathcal{D}$, whose positive set is $\mathcal{D}_c = \{z \in \mathcal{D} : I_c(z) = 1\}$. A model $\phi_\theta$ induces the \emph{concept geometry} 
\[
\conceptgeo_c^\theta = \phi_\theta(\mathcal{D}_c) \subset \mathbb{R}^d.
\]
\end{definition}

Fundamentally, a concept geometry is a set of points in the representation space, and characterizing the geometric structure of this point set is a central question in concept-based mechanistic interpretability. We approach this question by fitting a simple parametric shape to the point cloud, so that studying the concept geometry reduces to reading and tracking the fitted parameters.

\begin{definition}[Capsule]
\label{def:capsule}
Given a unit axis $a \in \mathbb{R}^d$ with $\|a\| = 1$, an axis span $s_a \in [0, 2]$, and norm bounds $0 \leq \nlow \leq \nhigh$, the corresponding \emph{capsule} is the region in the representation space:
\begin{equation}
    \capsule{a}{s_a}{\nlow}{\nhigh} = \left\{ x \in \mathbb{R}^d \;\middle|\; \dcos(x, a) \leq s_a,\; \nlow \leq \|x\| \leq \nhigh \right\}.
\end{equation}
\end{definition}

Geometrically, a capsule is an angular cone about the axis $a$ intersected with a norm shell. The motivation for using this form is that the aspects of concept geometry we care about can be directly read from the parameters of $\capsule{a}{s_a}{\nlow}{\nhigh}$. Specifically, the axis $a$ gives the central direction of the concept's representations; the axis span $s_a$ gives their angular spread around that direction; and the norm bounds $\nlow, \nhigh$ give the magnitude interval they occupy, whose width we write as the \emph{norm band} $\normband = \nhigh - \nlow$.

\subsection{Capsule Matching and Coverage Curves}
\label{sec:matching}

A concept geometry $\conceptgeo_c^\theta$ is in general an intractable point set. Prior work approximates such geometries as high-dimensional ellipsoids \citep{sorscher2022neural} or tracks their topology via persistent homology \citep{malhotra2026trackingrepresentationdynamicslarge}; we instead approximate $\conceptgeo_c^\theta$ with a finite sample of points, $\bar{\conceptgeo}_c^\theta = \{x_1, \ldots, x_m\} \subset \mathbb{R}^d$, extracted from $\phi_\theta$ on the positive set $\mathcal{D}_c$, and match a capsule to these points via the closed-form procedure of Algorithm~\ref{alg:matching}. Writing $\hat{x}_i = x_i / \|x_i\|$ for the unit-normalized samples, the axis is the normalized mean direction,
\begin{equation}
    a = \bar{r} / \|\bar{r}\|, \qquad \bar{r} = \frac{1}{m} \sum_{i=1}^{m} \hat{x}_i,
\end{equation}
the axis span is the maximum cosine distance from the samples to the axis, $s_a = \max_i \dcos(x_i, a)$, and the norm bounds are the minimum and maximum sample norms, $\nlow = \min_i \|x_i\|$ and $\nhigh = \max_i \|x_i\|$. The resulting capsule $\capsule{a}{s_a}{\nlow}{\nhigh}$ covers all fitting samples by construction and provides a compact, closed-form summary of the concept's geometry.

\begin{algorithm}[H]
\caption{Capsule Matching}
\label{alg:matching}
\begin{algorithmic}[1]
\REQUIRE samples $\bar{\conceptgeo}_c^\theta = \{x_1, \ldots, x_m\}$
\ENSURE capsule $\capsule{a}{s_a}{\nlow}{\nhigh}$
\STATE $\hat{x}_i \leftarrow x_i / \|x_i\|$ for $i = 1, \ldots, m$
\STATE $\bar{r} \leftarrow \frac{1}{m} \sum_{i=1}^{m} \hat{x}_i$; \quad $a \leftarrow \bar{r} / \|\bar{r}\|$ \hfill $\triangleright$ axis
\STATE $s_a \leftarrow \max_i \dcos(x_i, a)$ \hfill $\triangleright$ axis span
\STATE $\nlow \leftarrow \min_i \|x_i\|$; \quad $\nhigh \leftarrow \max_i \|x_i\|$ \hfill $\triangleright$ norm bounds
\RETURN $\capsule{a}{s_a}{\nlow}{\nhigh}$
\end{algorithmic}
\end{algorithm}

\paragraph{Coverage curves.} The capsule parameters summarize the boundary of the region a concept occupies, but miss how the concept's representations are distributed within it. We therefore introduce two coverage curves for the span parameter and the norm parameters: the span curve $F_s(t)$ and the norm curve $F_n(t)$. The \emph{span curve} of a capsule is the function
\begin{equation}
    F_s(t) = \frac{1}{\left| \bar{\conceptgeo}_c^\theta \right|} \left| \left\{ x_i \in \bar{\conceptgeo}_c^\theta \;\middle|\; 0 \leq \dcos(x_i, a) \leq t \right\} \right|,
\end{equation}
defined for $t \in [0, s_a]$, reporting the fraction of the concept's samples covered by the cone of axis $a$ and span $t$ as $t$ grows from $0$ to the fitted span. The \emph{norm curve} is defined analogously,
\begin{equation}
    F_n(t) = \frac{1}{\left| \bar{\conceptgeo}_c^\theta \right|} \left| \left\{ x_i \in \bar{\conceptgeo}_c^\theta \;\middle|\; \nlow \leq \|x_i\| \leq t \right\} \right|,
\end{equation}
defined for $t \in [\nlow, \nhigh]$, reporting the fraction of samples whose norms fall below $t$ within the fitted norm bounds. Both curves are empirical cumulative distributions of the quantities bounded by the capsule, increasing the resolution of \CapsuleLens{} beyond its boundary parameters: a span curve that rises steeply near \(0\) indicates representations concentrated tightly around the axis with the fitted span determined by a few outlying samples; a curve that rises late indicates a genuinely broad angular distribution; and flat segments separating steep rises suggest multiple angular clusters.

\subsection{Capsule Validation}
\label{sec:validation-protocol}

We validate a fitted capsule along two complementary dimensions: \emph{coverage} and \emph{disentanglement}. We split the positive set $\mathcal{D}_c$ into disjoint fitting and validation sets $\mathcal{D}_c^{\mathrm{fit}}$ and $\mathcal{D}_c^{\mathrm{val}}$, fit a capsule $\mathcal{C}_c$ on $\mathcal{D}_c^{\mathrm{fit}}$, and define its held-out coverage as
\begin{equation}
    \mathrm{Cov}(\mathcal{C}_c) = P(\phi(x)\in\mathcal{C}_c\mid I_c(x)=1) = \frac{1}{\left|\mathcal{D}_c^{\mathrm{val}}\right|}\left|\left\{z\in\mathcal{D}_c^{\mathrm{val}}\;\middle|\;\phi(z)\in\mathcal{C}_c\right\}\right|.
\end{equation}
This measures whether the fitted region generalizes to unseen samples of the same concept. To characterize its overlap with other concepts, we construct a negative validation set $\mathcal{D}_{\mathrm{not}\text{-}c}^{\mathrm{val}}$ by sampling examples that do not belong to concept $c$, and define the \emph{disentanglement score}
\begin{equation}
    \mathrm{Dis}(c)=P(I_c(x)=1\mid \phi(x)\in\mathcal{C}_c)=\frac{\rho^+}{\rho^++\rho^-},
\end{equation}
where $\rho^+=\frac{1}{|\mathcal{D}_c^{\mathrm{val}}|}\left|\left\{z\in\mathcal{D}_c^{\mathrm{val}}\;\middle|\;\phi(z)\in\mathcal{C}_c\right\}\right|$ and $\rho^-=\frac{1}{|\mathcal{D}_{\mathrm{not}\text{-}c}^{\mathrm{val}}|}\left|\left\{z\in\mathcal{D}_{\mathrm{not}\text{-}c}^{\mathrm{val}}\;\middle|\;\phi(z)\in\mathcal{C}_c\right\}\right|$. Higher $\mathrm{Dis}(c)$ indicates stronger separation of the target concept from others within the region $\mathcal{C}_c$.

\subsection{Tracking Concept Geometry}
\label{sec:tracking}

Because capsules are fitted independently at every model state, comparing the capsule $\mathcal{C}_c^{(t)}$ fitted at training checkpoint $t$ against the base capsule $\mathcal{C}_c^{(0)}$ provides a direct summary of how concept geometries change during training. Table~\ref{tab:drift-metrics} summarizes the drift metrics we track per concept and checkpoint. Aggregating these signals across concepts and layers turns a single capsule fit into a trajectory that localizes where training reshapes representations (by layer), and identifies which concepts it affects most (by concept). Section~\ref{sec:drift} demonstrates these can track representation drifts.

\begin{table}[t]
\caption{\textbf{Capsule drift metrics for tracking concept geometry across checkpoints.} Each metric compares the capsule fitted at checkpoint \(t\) against the base capsule at checkpoint \(0\). The axis drift \(\Delta a\) measures the rotation of the concept axis; the span change \(\Delta s_a\) measures the widening (\(>0\)) or narrowing (\(<0\)) of the concept's angular spread; and the norm-band changes \(\Delta \normband\), \(\Delta \nhigh\), \(\Delta \nlow\) measure how the magnitude interval the concept occupies expands, contracts, or shifts.}
\label{tab:drift-metrics}
\begin{center}
\begin{tabular}{lll}
\toprule
\textbf{Metric} & \textbf{Definition} & \textbf{Interpretation} \\
\midrule
\(\Delta a\) & \(\dcos(a^{(0)}, a^{(t)})\) & rotation of the concept axis \\
\(\Delta s_a\) & \(s_a^{(t)} - s_a^{(0)}\) & widening (\(>0\)) or narrowing (\(<0\)) of angular spread \\
\(\Delta \normband\) & \(\normband^{(t)} - \normband^{(0)}\) & expansion or contraction of the norm band \\
\(\Delta \nhigh\) & \(\nhigh^{(t)} - \nhigh^{(0)}\) & shift of the upper norm bound \\
\(\Delta \nlow\) & \(\nlow^{(t)} - \nlow^{(0)}\) & shift of the lower norm bound \\
\bottomrule
\end{tabular}
\end{center}
\end{table}

\section{Validating Capsule Lens}
\label{sec:validation}

We validate the reliability of \CapsuleLens{} for locating and tracking concept geometry. Section~\ref{subsec:concepts} describes the construction of visual and textual concepts used throughout our experiments. Section~\ref{sec:validation-results} evaluates fitted capsules using held-out coverage and disentanglement metrics. Section~\ref{sec:geometry-shapes} compares the capsule with alternative geometric approximations. Further robustness analyses on sampling and boundary estimation are provided in Appendix~\ref{app:validation}.

\subsection{Concept Construction}
\label{subsec:concepts}

We construct both visual and textual concepts, where each concept is represented by a collection of samples exhibiting certain property. These collections of data samples can be built with human annotation or as token groups in context.

\paragraph{Visual concepts.} We construct visual concepts from Visual Genome~\citep{krishna2017visualgenome}, where human annotators provide named bounding boxes for objects and regions. An image is considered positive for concept $c$ if it contains a bounding box labelled $c$ covering at least $1\%$ of the image. For each concept, we retain the $m=100$ images in which the corresponding bounding box occupies the largest fraction. Across the resulting $1{,}000$ concepts, the median bounding-box area is $30\%$ of the image. The samples for each concept are then divided into disjoint fitting and validation sets.

\paragraph{Textual concepts.} We construct textual concepts in two complementary ways, either stated as \textbf{token groups in context} or \textbf{labeled data samples with human annotations}. A \emph{lexical} concept is defined by grouping occurrences of one or more related tokens across different contexts. We construct $18$ lexical concepts from GSM8K~\citep{cobbe2021gsm8k}, covering arithmetic operations, units of measurement, and recurring entities, and $1{,}026$ lexical concepts from the Brown~\citep{francis1964brown}, NLTK Gutenberg~\citep{bird2009natural}, and Reuters RCV1~\citep{lewis2004rcv1} corpora, spanning general-domain language. A \emph{labelled} concept is defined by a human-annotated semantic category. We construct $43$ labelled concepts from SemCor~\citep{miller1993semcor}. Each sample is represented by the token sequence cropped at the corresponding token position.

\subsection{Held-Out Coverage and Disentanglement of Capsules}
\label{sec:validation-results}

We evaluate held-out coverage and disentanglement across all model components and concept types introduced above. For each concept, we use $50$ samples for capsule matching and a disjoint set of $50$ samples for held-out coverage. Disentanglement is evaluated using balanced positive and negative samples, $|\mathcal{D}_{c}^{\mathrm{val}}|=|\mathcal{D}_{\mathrm{not}\text{-}c}^{\mathrm{val}}|$. Table~\ref{tab:validation-main} summarizes the results.

Fitted capsules achieve consistently high held-out coverage across models, components, and concept types, with an overall mean of $0.943$. Disentanglement exhibits substantially greater variation: visual concepts remain close to the balanced baseline across most components, whereas textual concepts are more strongly separated, particularly lexical concepts.

\begin{table}[h]
\centering
\caption{\textbf{Held-out coverage and disentanglement across models and components.} We report the mean and standard deviation of $\mathrm{Cov}$ and $\mathrm{Dis}$ across fitted capsules.}
\vspace{2mm}
\label{tab:validation-main}
\small
\setlength{\tabcolsep}{3.5pt}
\begin{tabular}{lllrrrrrr}
\toprule
Model & Component & Type & Layers & Capsules & Cov. & Std. & Dis. & Std. \\
\midrule
CLIP ViT-B/32   & image enc.  & visual   & 13 & 13{,}000 & 0.941 & 0.048 & 0.551 & 0.093 \\
CLIP ViT-L/14   & image enc.  & visual   & 25 & 25{,}000 & 0.941 & 0.048 & 0.533 & 0.073 \\
\midrule
Qwen2-VL-2B     & ViT         & visual   & 32 & 32{,}000 & 0.946 & 0.045 & 0.504 & 0.013 \\
Qwen2-VL-2B     & LLM         & visual   & 28 & 28{,}000 & 0.947 & 0.044 & 0.507 & 0.016 \\
\midrule
LLaVA-1.5-7B    & ViT (patch) & visual   & 24 & 24{,}000 & 0.934 & 0.051 & 0.506 & 0.015 \\
LLaVA-1.5-7B    & ViT (CLS)   & visual   & 24 & 24{,}000 & 0.941 & 0.047 & 0.530 & 0.070 \\
LLaVA-1.5-7B    & projector   & visual   &  1 &  1{,}000 & 0.943 & 0.047 & 0.503 & 0.011 \\
LLaVA-1.5-7B    & LLM         & visual   & 32 & 32{,}000 & 0.941 & 0.048 & 0.505 & 0.013 \\
\midrule
Qwen2.5-1.5B    & LLM         & lexical  & 28 & 28{,}000 & 0.950 & 0.048 & 0.786 & 0.225 \\
Qwen2.5-1.5B    & LLM         & labelled & 28 &  1{,}204 & 0.937 & 0.052 & 0.566 & 0.084 \\
Qwen2.5-1.5B    & LLM         & math     & 28 &    504 & 0.932 & 0.053 & 0.607 & 0.126 \\
\midrule
Qwen2.5-14B     & LLM         & lexical  & 48 & 48{,}000 & 0.947 & 0.047 & -- & -- \\
Qwen2.5-14B     & LLM         & labelled & 48 &  2{,}064 & 0.934 & 0.053 & -- & -- \\
Qwen2.5-14B     & LLM         & math     & 48 &    864 & 0.929 & 0.062 & -- & -- \\
\midrule
All             &             &          & 407 & 259{,}636 & 0.943 & 0.048 & -- & -- \\
\bottomrule
\end{tabular}
\end{table}

\subsection{Comparison with Alternative Geometric Approximations}
\label{sec:geometry-shapes}

We compare the capsule with four alternative geometric approximations---a cone, ball, norm band, and axis-aligned box---fitted to the same concept representations. We evaluate each choice using held-out coverage and disentanglement, together with the geometric information that can be directly extracted from its parameters, including angular structure, norm structure, and whether the two can be separated. Table~\ref{tab:geometry-shapes} summarizes these comparisons under the original min/max boundary.

Capsules achieve comparable held-out coverage to cones and balls while attaining the highest disentanglement among the geometric approximations with meaningful coverage. Moreover, the capsule provides a more informative description of concept geometry by separating angular extent from representation norm. We therefore adopt it as a compact approximation whose parameters directly characterize distinct geometric properties of the concept region.

\vspace{-3mm}
\begin{table}[h]
\centering
\caption{\textbf{Comparison with alternative geometric approximations.} Held-out coverage and disentanglement $\mathrm{Dis}$ under the original min/max boundary ($q=100$), averaged across concept types, model components, and representative layers. $^{*}$The box exhibits extremely low held-out coverage because coordinate-wise bounds generalize poorly in high-dimensional anisotropic representation spaces, a manifestation of the curse of dimensionality~\citep{koppen2000curse}.}
\label{tab:geometry-shapes}
\small
\setlength{\tabcolsep}{4pt}
\begin{tabular}{lrrrccc}
\toprule
Shape & Params & Coverage & $\mathrm{Dis}$ & Angular & Norm & Separable \\
\midrule
Capsule & $d_{\mathrm{rep}}+3$ & 0.943 & 0.583 & \checkmark & \checkmark & \checkmark \\
Cone    & $d_{\mathrm{rep}}+1$ & 0.975 & 0.578 & \checkmark & $\times$ & $\times$ \\
Ball    & $d_{\mathrm{rep}}+1$ & 0.970 & 0.568 & \checkmark & \checkmark & $\times$ \\
Band    & $2$                  & 0.962 & 0.514 & $\times$ & \checkmark & $\times$ \\
Box     & $2d_{\mathrm{rep}}$  & $0.012^{*}$ & 0.653 & \checkmark & \checkmark & $\times$ \\
\bottomrule
\end{tabular}
\end{table}
\vspace{-3mm}

\section{Analyzing Concept Geometry in Static Representations}
\label{sec:static}
We next use \CapsuleLens{} to analyze concept geometry within static representations. Section~\ref{sec:static-depth} examines the internal structure of concept geometry. Section~\ref{sec:disentanglement-depth} shows how \CapsuleLens{} can uncover geometric differences across model layers.

\subsection{Internal Structure of Concept Geometry}
\label{sec:static-depth}

\begin{figure}[h]
    \centering
    \includegraphics[width=0.73\linewidth]{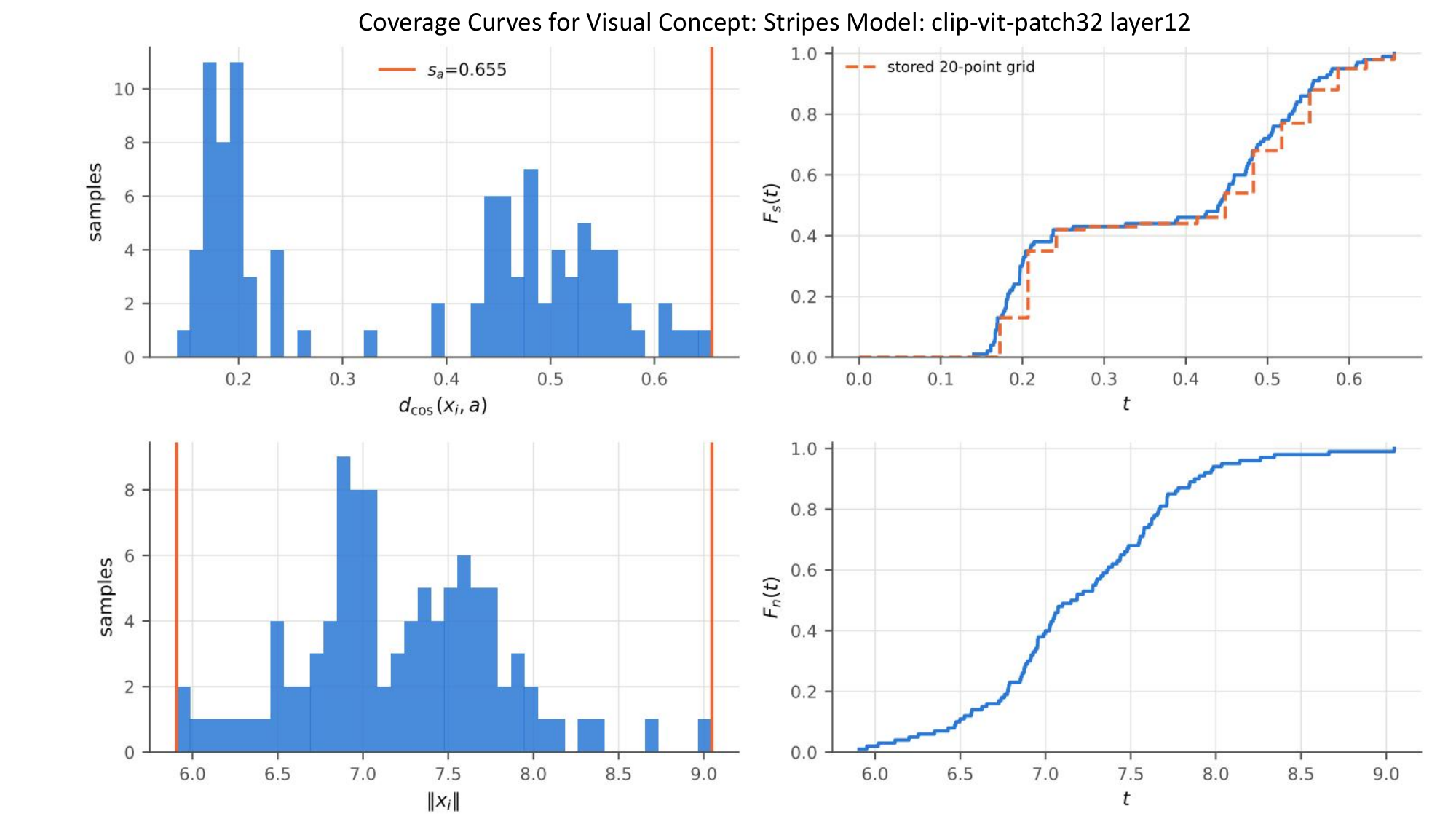}
    \vspace{-5mm}
    \caption{\textbf{Coverage curves reveal internal concept geometry beyond capsule boundaries.} For \emph{stripes} at layer 12 of CLIP ViT-B/32, the span distribution and $F_s(t)$ reveal two angular clusters, while the norm distribution and $F_n(t)$ vary smoothly. The fitted span $s_a$ captures the outer extent of the concept region, whereas the coverage curves reveal how representations populate that region.}
    \label{fig:stripes-detail}
\end{figure}

The capsule parameters describe the boundary of a concept's representation region, while the span curve $F_s(t)$ and norm curve $F_n(t)$ reveal how representations are distributed inside that boundary. Figure~\ref{fig:stripes-detail} illustrates this distinction for the concept \emph{stripes} at layer 12 of CLIP ViT-B/32. Its span curve exhibits two sharp rises separated by a plateau, revealing two angular clusters that cannot be inferred from the fitted span $s_a=0.655$ alone. In contrast, the norm curve rises smoothly, indicating a comparatively unimodal distribution in representation magnitude. Thus, concepts with similar capsule boundaries can exhibit substantially different internal geometric structure, which the coverage curves make explicit.

\subsection{Layer-wise Evolution of Concept Disentanglement}
\label{sec:disentanglement-depth}

We examine how concept disentanglement changes across model depth. For visual concepts, we compute the mean disentanglement score $\mathrm{Dis}$ at each relative layer depth across CLIP, Qwen2-VL, and LLaVA components; for textual concepts, we report the corresponding layer-wise scores for lexical, math, and labelled concepts in Qwen2.5-1.5B.

\begin{figure}[H]
    \centering
    \includegraphics[width=\linewidth]{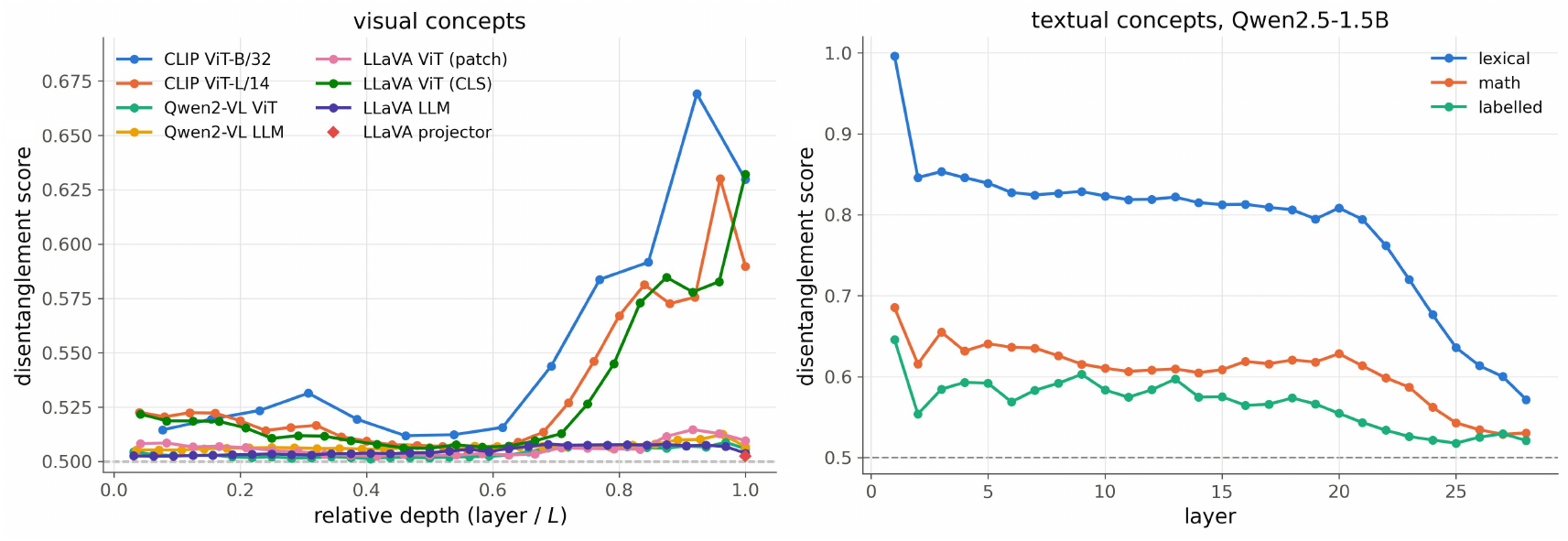}
    \vspace{-8mm}
    \caption{\textbf{Concept disentanglement across model depth.} \textbf{Left:} visual concepts across CLIP, Qwen2-VL, and LLaVA components, showing pronounced late-layer increases in disentanglement for several visual representations. \textbf{Right:} textual concepts in Qwen2.5-1.5B, where disentanglement generally decreases toward later layers, most prominently for lexical concepts.}
    \label{fig:disentanglement}
\end{figure}

Figure~\ref{fig:disentanglement} reveals contrasting depth profiles across modalities. Several visual encoders show a pronounced increase in disentanglement toward their final layers, consistent with geometric separation phenomena such as neural collapse~\citep{Papyan_2020}. In contrast, textual concepts become progressively more entangled in later LLM layers, particularly for lexical concepts, consistent with prior findings that lexical identity and lexical semantics are strongest in earlier representations and become less explicit with depth~\citep{liu2024fantasticsemanticstheminvestigating,li2026modelinternalsleuthingfinding}.

\section{Tracking Concept Geometry in Dynamic Representations}
\label{sec:drift}
We demonstrate how Capsule Lens tracks concept geometry drifts induced by different training pipelines. Section~\ref{sec:drift-pretraining} studies CLIP pretraining, Section~\ref{sec:drift-rl} examines RL post-training on visual question answering, and Section~\ref{sec:drift-case3} analyzes RL post-training on mathematical reasoning.

\subsection{Case Study I: CLIP Pretraining}
\label{sec:drift-pretraining}

We first study how concept geometry forms during pretraining from scratch. We train a standard CLIP model~\citep{radford2021learningtransferablevisualmodels} with a symmetric InfoNCE objective on MSCOCO~\citep{lin2015microsoftcococommonobjects}, CC3M~\citep{sharma2018conceptual}, and CC12M~\citep{changpinyo2021conceptual12mpushingwebscale}. We track the visual concepts of Section~\ref{subsec:concepts} at five ViT blocks and six training checkpoints, using epoch $10$ as the reference because the randomly initialized model has no meaningful concept axes.

\begin{figure}[h]
    \centering
    \includegraphics[width=0.75\linewidth]{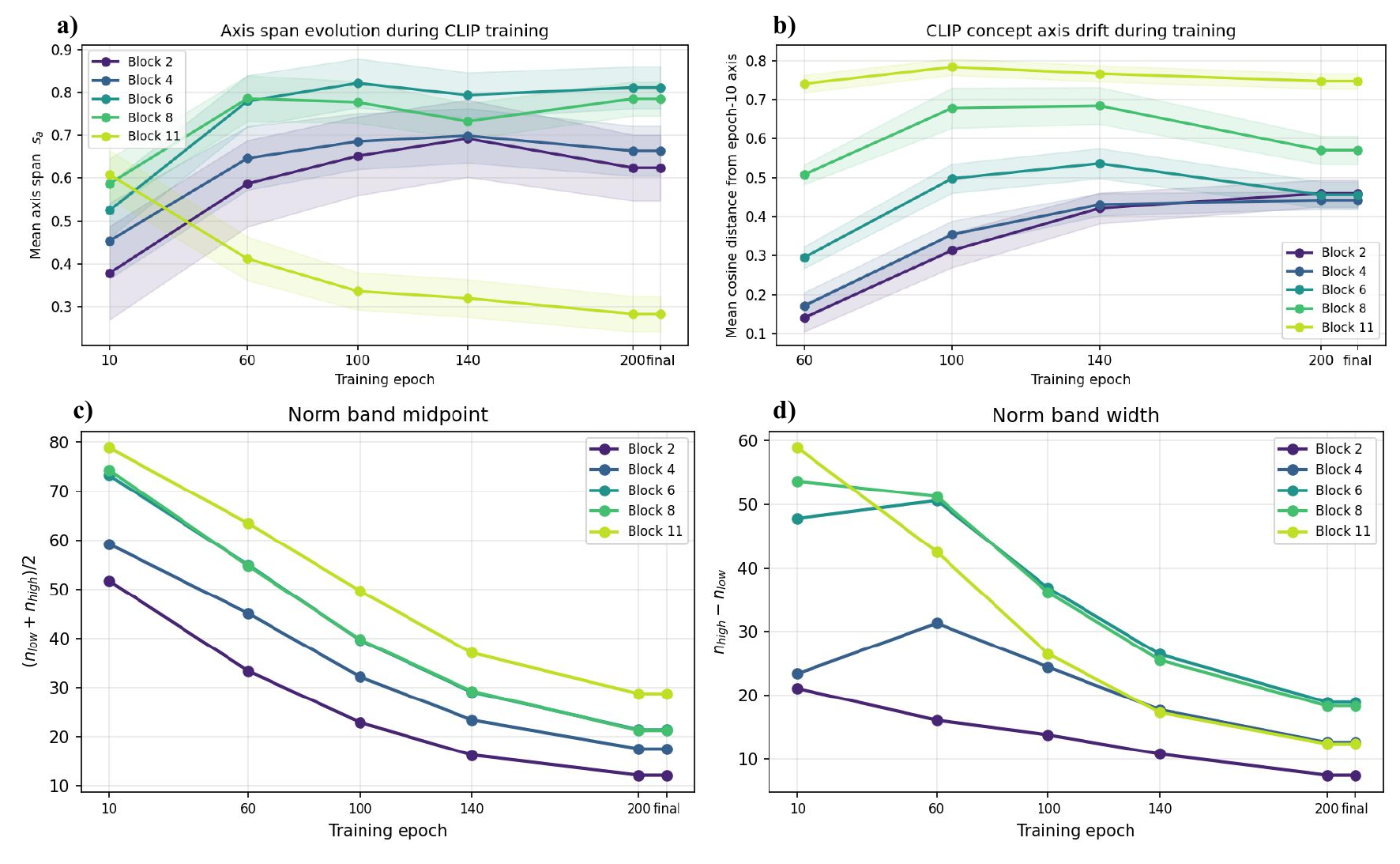}
    \vspace{-6mm}
    \caption{\textbf{CLIP pretraining restructures concept geometry throughout the network.} \textbf{a)} Axis spans widen in blocks $2$--$8$ but contract strongly in the final block. \textbf{b)} Concept axes rotate substantially at all probed depth. \textbf{c--d)} Norm-band midpoint and width contract steadily throughout training.}
    \label{fig:clip-pretraining}
\end{figure}

Pretraining substantially restructures concept geometry throughout the network (Figure~\ref{fig:clip-pretraining}). Concept axes rotate at every probed block, while span dynamics differ sharply with depth: blocks $2$--$8$ widen whereas the final block contracts from $s_a\approx0.61$ to $0.28$. Norm bands also contract steadily across depth. Per-concept drift is comparatively uniform, indicating that pretraining produces broad rather than strongly concept-selective geometric restructuring. Appendix~\ref{app:clip-pretraining} shows additional analyses.

\subsection{Case Study II: RL Post-Training on Visual Question Answering}
\label{sec:drift-rl}

We next study RL post-training on VLMs (Appendix~\ref{sec:app-exp2}). We fine-tune Qwen2-VL~\citep{qwen2} with PPO and LoRA on OKVQA~\citep{marino2019okvqavisualquestionanswering}, tracking the visual concepts of Section~\ref{subsec:concepts} at five layers and five checkpoints. Capsules from the base model serve as the reference for all drifts.

\begin{figure}[h]
    \centering
    \includegraphics[width=0.75\linewidth]{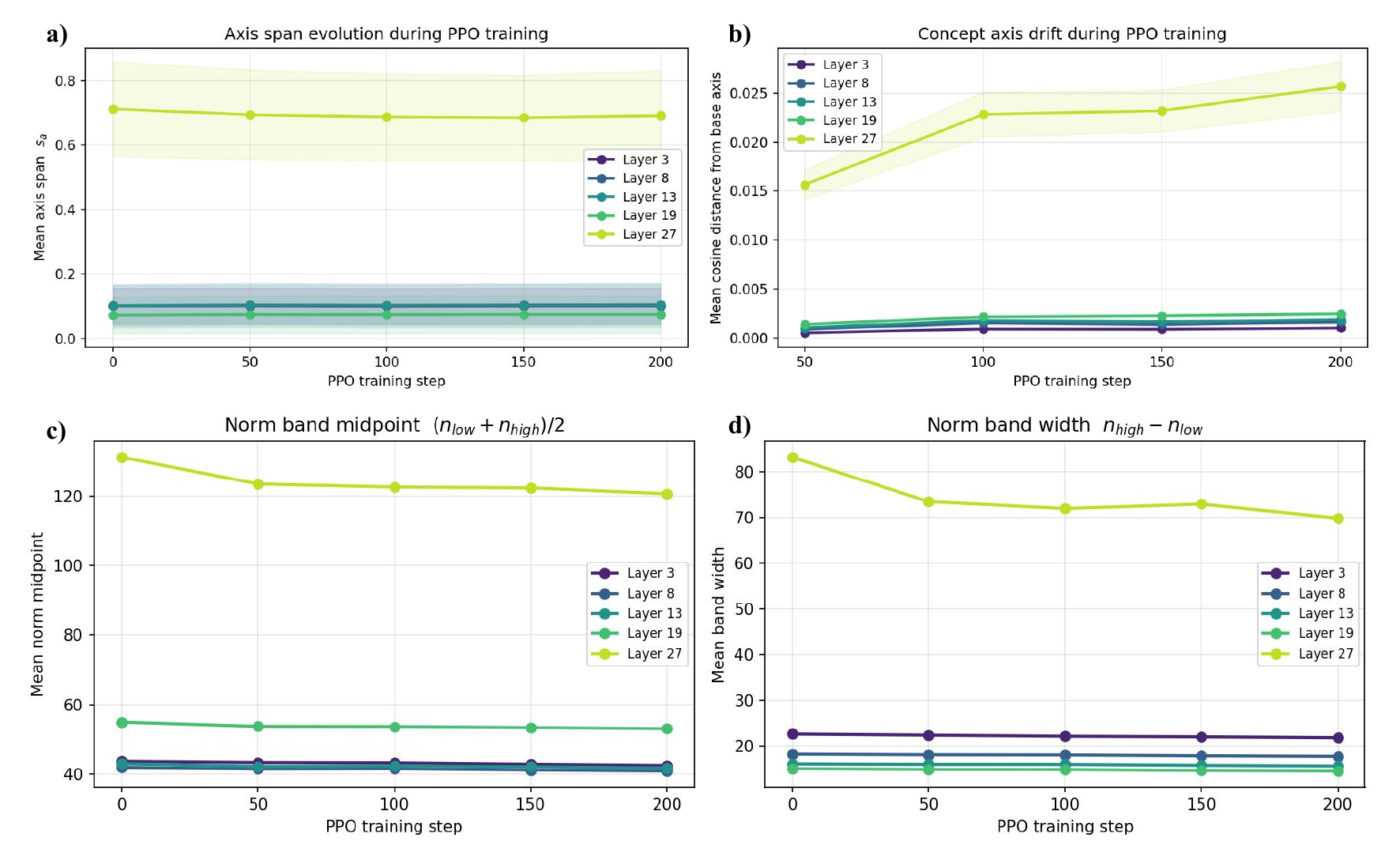}
    \vspace{-6mm}
    \caption{\textbf{PPO post-training concentrates geometric change in the final layer.} \textbf{a)} Axis spans remain nearly unchanged. \textbf{b)} Axis drift occurs predominantly at layer $27$. \textbf{c--d)} The final-layer norm band contracts rapidly and then stabilizes while axis rotation continues.}
    \label{fig:ppo-posttraining}
\end{figure}

Unlike pretraining, PPO-induced geometric change is highly localized in depth (Figure~\ref{fig:ppo-posttraining}). Axis spans remain nearly unchanged, while axis drift concentrates almost entirely in the final layer, reaching $\Delta a\approx0.026$ as earlier layers remain close to their base geometry. The final-layer norm band contracts rapidly within the first $50$ steps and then stabilizes even as its axis continues to rotate. PPO therefore induces substantially smaller and more localized geometric restructuring than pretraining.

\subsection{Case Study III: RL Post-Training on Mathematical Reasoning}
\label{sec:drift-case3}

Finally, we study RLVR for LLM reasoning (Appendix~\ref{app:case3}). We train Qwen2.5-1.5B with GRPO~\citep{shao2024deepseekmathpushinglimitsmathematical} on $500$ GSM8K problems~\citep{cobbe2021gsm8k} for $750$ optimizer steps, using answer correctness with an additional formatting bonus. After training, held-out GSM8K accuracy have moderate improvement from $0.512$ to $0.536$. We track the $18$ mathematical concepts of Section~\ref{subsec:concepts} at five layers and four checkpoints relative to the base model.

\begin{figure}[h]
    \centering
    \includegraphics[width=0.85\linewidth]{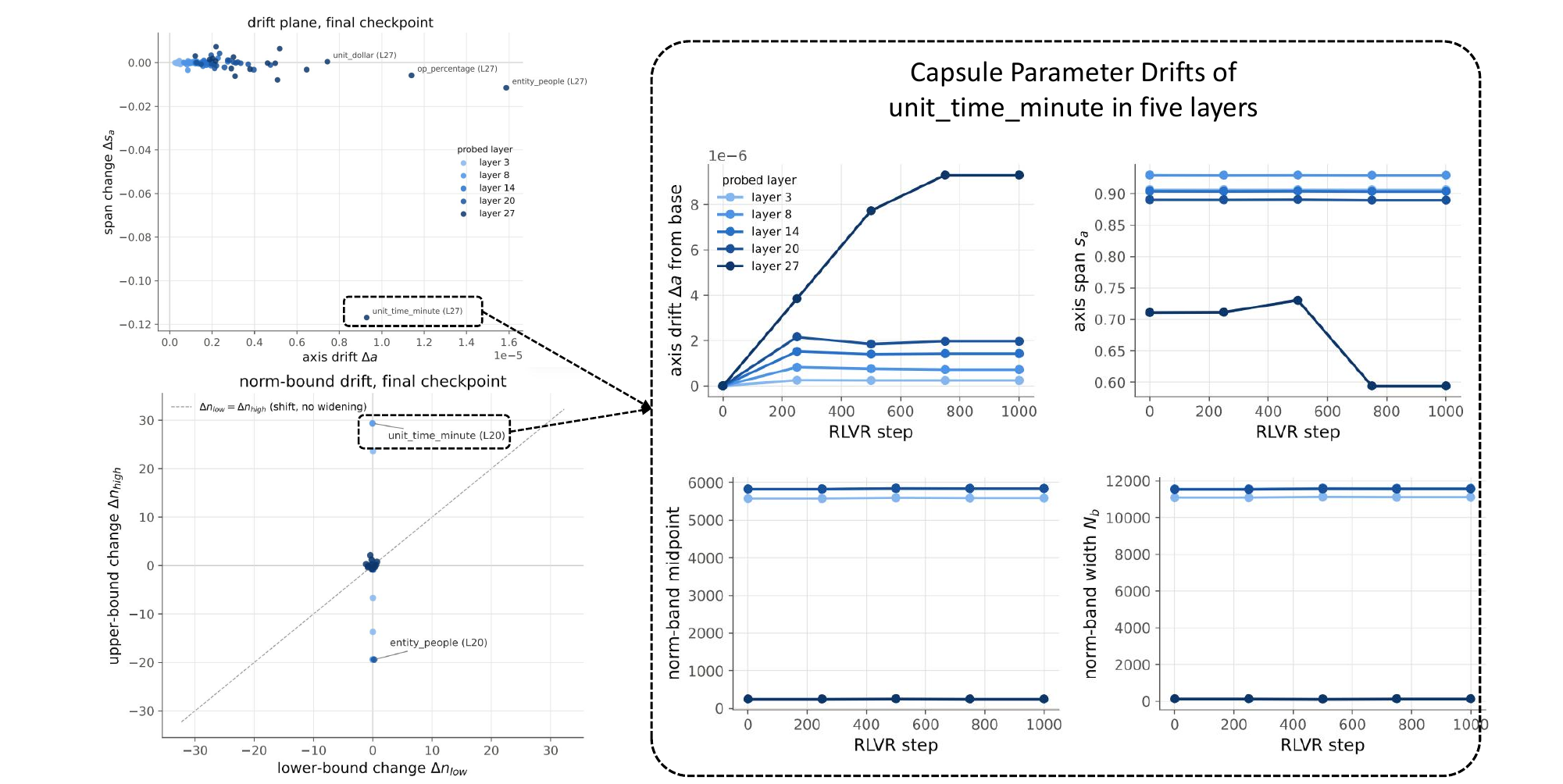}
    \vspace{-4mm}
    \caption{\textbf{RLVR induces sparse, concept-specific concentration of representation geometry.} \textbf{Left:} at the final checkpoint, most concept--layer pairs remain close to their base geometry, while a small number exhibit pronounced span contractions or norm-bound shifts. \textbf{Right:} for the representative concept \texttt{unit\_time\_minute}, the final-layer axis span contracts sharply late in training while norm geometry remain comparatively stable. These results show that RLVR selectively sharpens particular concept geometries at particular layers rather than comprehensively change the model.}
    \label{fig:rlvr-specificity}
\end{figure}

RLVR produces a third geometric pattern: changes are sparse across both concepts and layers. As shown in Figure~\ref{fig:rlvr-specificity}, most concept--layer pairs remain close to their base geometry, while a small subset exhibits pronounced span contractions or norm-bound shifts. The clearest example is \texttt{unit\_time\_minute}, whose layer-$27$ span contracts from $s_a=0.711$ to $0.594$ late in training while its axis orientation and norm geometry remain comparatively stable. Thus, unlike the broad restructuring of pretraining, RLVR selectively concentrates particular concept geometries at particular depths while producing a moderate improvement in held-out task accuracy.

\vspace{-2mm}

\section{Conclusion}
\label{sec:conclusion}

In this work, we introduce Capsule Lens, a novel framework that can locate and characterize concept geometry in static representations and can track how that geometry changes in dynamic representations. We validate Capsule Lens systematically and show how it can be used to analyze concept geometry across a range of models. We further demonstrate, through three case studies spanning distinct training pipelines, how Capsule Lens can track concept geometry as it shifts under training.

\newpage

\subsection*{AI Use Statement}

In this work, we used generative AI tools, including ChatGPT, Claude, and Claude Code to implement methods, write and edit code, and assist in drafting the manuscript. We did not use generative AI tools to generate synthetic data, formulate or prove mathematical claims, or propose our core hypotheses, methodology, or experimental design; these were carried out by the authors. We have reviewed all AI-assisted work: code was checked against expected behavior and the underlying data, and all AI-drafted text and result interpretations were verified against experimental outputs and revised where necessary. We take responsibility for the final content of this work, including text, claims, and artifacts produced with the aid of generative AI.

\subsection*{Ethics Statement}

This work uses only publicly available, non-sensitive data and involves no human-subjects research, personal data, or dual-use capability beyond that of the underlying pretrained models we analyze. We do not foresee ethics concerns specific to this submission.

\subsection*{Reproducibility Statement}

\CapsuleLens{} is fully specified in closed form: the capsule-fitting procedure and the associated span and norm coverage curves are detailed in Sec.~\ref{sec:method}, and the validation protocol (held-out coverage, concept-specificity, and sampling-stability checks) is described in Sec.~\ref{sec:method} and Sec.~\ref{sec:validation}. All concept construction, model choices, and training details for the case studies are provided in the main text and appendices. Code implementing capsule fitting, validation, and all reported experiments will be released upon publication.

\bibliography{iclr2027_conference}
\bibliographystyle{iclr2027_conference}

\appendix

\newpage
\section{Related Work}

\subsection{Lens to Interpret Model Internals}
Diverse methods have been proposed to interpret a model's internal representation. The Logit Lens \citep{nostalgebraist2020logitlens} projects intermediate hidden states through the unembedding to reveal a layerwise prediction trajectory, and the Tuned Lens \citep{belrose2025elicitinglatentpredictionstransformers} corrects its basis-mismatch failures with learned per-layer translators. Patchscopes \citep{ghandeharioun2024patchscopesunifyingframeworkinspecting} patches hidden states into inspection prompts to decode them in natural language, and Rep2Text \citep{zhao2025rep2text} decodes full text from a single token representation. Probing classifiers \citep{alain2018understandingintermediatelayersusing, hewitt-manning-2019-structural} test whether predefined concepts are linearly decodable. Sparse autoencoders decompose polysemantic activations into dictionaries of monosemantic features \citep{cunningham2023sparseautoencodershighlyinterpretable, gao2024scalingevaluatingsparseautoencoders, bussmann2025learningmultilevelfeaturesmatryoshka}, with LanSE grounding features directly in natural language \citep{tang2025human} and natural language autoencoders producing unsupervised explanations of activations \citep{fraser2026natural}. For comparison across models and training, crosscoders diff models through shared dictionaries \citep{lindsey2024sparse, jiralerspong2026crossarchitecturemodeldiffingcrosscoders} and SAE Track follows feature dynamics along a training trajectory \citep{xu2025trackingfeaturedynamicsllm}. Despite their diversity, all of these lenses map activations onto tokens, language, or predefined and learned features, spaces that are easier to interpret, but none directly reports \emph{how concepts are encoded} in representation space---the spread, anisotropy, and norm structure of the region a concept occupies.

\subsection{Neural Geometry}
Neural geometry refers to the study of how information is spatially organized in the representation spaces of neural networks, often through overarching hypotheses supported by suggestive yet partial evidence. The linear representation hypothesis holds that high-level concepts are encoded as directions \citep{park2024linearrepresentationhypothesisgeometry}, with roots in the linear analogies of word embeddings \citep{korchinski2025emergencelinearanalogiesword}; superposition further explains how more features than dimensions can coexist as almost-orthogonal directions \citep{elhage2022toymodelssuperposition}, and diverse evidence supports this linear view, from refusal directions to linear representations of space and time and emergent world models \citep{arditi2024refusallanguagemodelsmediated, gurnee2024languagemodelsrepresentspace, li2024emergentworldrepresentationsexploring}. Beyond directions, various works uncover richer, nonlinear structures: categorical and hierarchical concepts form polytopes and orthogonal hierarchies \citep{park2025geometrycategoricalhierarchicalconcepts}, some features are irreducibly multi-dimensional, forming circles for days and months \citep{engels2025languagemodelfeaturesonedimensionally}, SAE feature clouds exhibit multi-scale structure \citep{Li_2025}, and manifold steering reveals shared geometry between representation and behavior \citep{wurgaft2026manifoldsteeringrevealsshared}. At a global level, neural collapse suggests that class representations might collapse to a simplex structure in the terminal phase of training \citep{Papyan_2020}, while the platonic representation hypothesis suggests that representations across models and modalities might converge \citep{huh2024platonicrepresentationhypothesis}. However, we currently lack a method to track how training reshapes neural geometry; such changes remain characterized mostly by hypotheses that are not yet sufficiently supported.

\subsection{Understanding How Post-Training (SFT \& RL) Works}
Supervised fine-tuning (SFT) and reinforcement learning (RL) are the prevailing paradigms of model post-training, widely utilized for human alignment via RLHF and RLAIF \citep{ouyang2022traininglanguagemodelsfollow, lee2024rlaifvsrlhfscaling} and for enhancing reasoning via RLVR \citep{shao2024deepseekmathpushinglimitsmathematical, wang2025reinforcementlearningreasoninglarge}. Yet how these methods actually work remains poorly understood, and existing works hold seemingly contradictory views. Some argue RLVR merely sharpens the output distribution over solutions the base model can already produce \citep{yue2025doesreinforcementlearningreally, shao2026spuriousrewardsrethinkingtraining} and may even collapse the model's capability boundary \citep{dong2026rlpluscounteringcapabilityboundary}; others find RL genuinely incentivizes correct reasoning \citep{wen2025reinforcementlearningverifiablerewards} and that prolonged RL expands reasoning boundaries toward novel capabilities \citep{liu2025prorlprolongedreinforcementlearning, shen2026doesrlvrextendreasoning}. Comparative studies report that SFT tends to memorize training data while RL generalizes to unseen rule and visual variants, in both LLMs and VLMs \citep{chu2025sftmemorizesrlgeneralizes}. However, others show this narrative is conditional rather than universal, with SFT's generalization depending on optimization choices, data composition, and model capability \citep{ren2026rethinkinggeneralizationreasoningsft}. Finer-grained analyses identify key components of RL, such as high-entropy forking tokens \citep{wang20258020rulehighentropyminority}, and show negative reinforcement mitigates diversity loss \citep{zhu2025surprisingeffectivenessnegativereinforcement}. Crucially, almost all of these works provide perspectives at the behavior level, measuring shifts in the output distributions, and do not answer what RL and SFT do to the representation space itself.

\newpage
\section{Discussion}
\label{sec:discussion}

\paragraph{Tracking representation geometry.} Mechanistic interpretability has largely focused on characterizing representations at a fixed model checkpoint, through directions, features, probes, or geometric structure \citep{park2024linearrepresentationhypothesisgeometry, engels2025languagemodelfeaturesonedimensionally, lindsey2024sparse, jiralerspong2026crossarchitecturemodeldiffingcrosscoders,202607.1441}. Comparatively less attention has been paid to how the geometry of those representations evolves throughout training, although recent work has begun to study feature and circuit dynamics across checkpoints \citep{xu2025trackingfeaturedynamicsllm, wang2025understandingfinetuningmechanismsllms}. This question is increasingly important as modern models undergo multiple stages of pretraining and post-training whose effects on internal representations remain only partially understood \citep{chu2025sftmemorizesrlgeneralizes, yue2025doesreinforcementlearningreally, wen2025reinforcementlearningverifiablerewards}. We therefore view tracking representation geometry as a complementary problem to interpreting a static representation: rather than asking only what a model represents, we also ask where and how that representation changes as the model is trained. Capsule Lens provides a simple way to make such changes measurable by matching the same interpretable geometric parameterization across model states.

\paragraph{Coverage and disentanglement as recall and precision.} The two metrics used in this work admit a natural interpretation analogous to recall and precision. Held-out coverage measures the fraction of unseen positive samples captured by a fitted capsule and can therefore be viewed as the \emph{recall} of the capsule in locating a concept's geometry. Conversely, the disentanglement score measures the probability that a representation admitted by a capsule belongs to its target concept and can be viewed as its \emph{precision}. This distinction is important because prior work suggests that concept representations need not be perfectly isolated or one-dimensional, but may instead occupy overlapping, multidimensional, or manifold-like regions \citep{sorscher2022neural, engels2025languagemodelfeaturesonedimensionally, modell2025originsrepresentationmanifoldslarge, bhalla2026sparseautoencoderscaptureconcept}. Since a capsule is an intentionally simple geometric family fitted in closed form, high held-out coverage indicates that the fitted region generalizes to unseen instances of the concept, whereas high disentanglement indicates that this simple region is relatively specific to that concept. We therefore view disentanglement not only as a property of the fitting procedure, but also as a characteristic of the underlying concept geometry: concepts whose representations are compactly approximated by a capsule should naturally achieve higher precision than concepts whose geometry is strongly intertwined with other concepts.

\paragraph{Why capsules?} Our choice of capsules is motivated primarily by simplicity and interpretability rather than by the claim that concept geometry must take this form. Prior work has described concept representations using directions \citep{park2024linearrepresentationhypothesisgeometry, arditi2024refusallanguagemodelsmediated}, higher-dimensional manifolds \citep{sorscher2022neural, engels2025languagemodelfeaturesonedimensionally, modell2025originsrepresentationmanifoldslarge}, categorical and hierarchical geometric structures \citep{park2025geometrycategoricalhierarchicalconcepts}, and cone-like regions \citep{wollschläger2026geometryrefusallargelanguage}. Capsules draw inspiration from these geometric views while retaining only a small number of directly interpretable parameters: the axis describes orientation, the axis span describes angular spread, and the norm bounds describe radial extent. These parameters can be computed in closed form and compared directly across layers, models, and training checkpoints. More expressive geometric descriptions, including ellipsoidal or topological characterizations \citep{sorscher2022neural, malhotra2026trackingrepresentationdynamicslarge}, may capture richer structure but are less immediately reducible to a small set of quantities whose changes can be interpreted across model states. Capsule Lens therefore deliberately trades geometric flexibility for a compact and trackable description of representation.

\paragraph{From case studies to systematic hypotheses.} The three dynamic case studies reveal several geometric patterns that merit substantially more systematic investigation. CLIP pretraining produces broad restructuring across the network, PPO post-training on visual question answering produces much smaller and strongly depth-localized changes, and RLVR on mathematical reasoning reveals sparse concept- and layer-specific concentration. These observations connect to a growing literature showing that pretraining, fine-tuning, and reinforcement learning can affect model representations in qualitatively different ways \citep{chu2025sftmemorizesrlgeneralizes, xu2025trackingfeaturedynamicsllm, liu2025prorlprolongedreinforcementlearning, yue2025doesreinforcementlearningreally, wen2025reinforcementlearningverifiablerewards, wang20258020rulehighentropyminority}. They also raise broader questions about which geometric changes are associated with behavioral improvements, whether different training objectives produce characteristic geometric signatures, and whether these patterns generalize across models, datasets, optimization settings, and random seeds. Establishing such general laws would require dedicated experimental studies for each training regime and is beyond the scope of the present work. Our goal here is instead to demonstrate that Capsule Lens provides a practical instrument for posing and studying these questions. We therefore treat the findings of our case studies as empirical hypotheses suggested by the framework and leave their systematic validation to future work.

\section{Future Work}
\label{sec:future-work}

We highlight four promising directions for future work:

\begin{itemize}
    \item \textbf{Systematically tracking specific training pipelines.} Capsule Lens can be used to study in detail how particular training procedures reshape concept geometry. Extending the case studies in this work into dedicated analyses across objectives, datasets, checkpoints, and model families can provide key insights to better understand different training pipelines.

    \item \textbf{Characterizing geometric differences across model layers.} Capsule Lens can be used to systematically compare concept geometry across depth, helping reveal how concepts are formed, transformed, concentrated, or disentangled between different layers of a model.

    \item \textbf{Studying hierarchical concepts through capsule intersections.} Capsule Lens may offer a geometric view of concept hierarchy by analyzing how capsules corresponding to related concepts intersect, overlap, or contain one another. This could provide a direct way to investigate how hierarchical semantic structure is organized in representation space.

    \item \textbf{Steering representations with capsules.} Beyond analysis, the geometric structure identified by Capsule Lens may provide a basis for representation steering. Interventions along a capsule's axis, within its angular span, or toward particular regions of its norm band could offer a geometrically grounded way to manipulate concept-related representations and study their causal effects on model behavior.
    
\end{itemize}
\newpage
\newpage
\section{Systematically Validating Capsule Lens}
\label{app:validation}

We provide systematic experiments further validating the reliability and design choices of \CapsuleLens{}. Section~\ref{sec:validation-samplesize} examines the effect of the number of samples used for capsule matching and the stability of the estimated geometry under sampling. Section~\ref{sec:validation-boundary} evaluates percentile-based capsule boundaries in place of the minimum and maximum fitting samples, examining the resulting trade-off between held-out coverage and disentanglement. Section~\ref{sec:validation-shapes} further compares capsules with alternative geometric approximations across different boundary choices.

\subsection{Effect of the Number of Samples per Concept}
\label{sec:validation-samplesize}

We examine how the number of samples used for capsule matching affects the estimated geometry. We vary $m$ from $5$ to $100$ while evaluating each fitted capsule on the same set of $100$ held-out samples. The experiment covers $796$ visual concepts across eight model components, using three representative layers from each component and three independent draws per concept.

\begin{figure}[h]
    \centering
    \includegraphics[width=\linewidth]{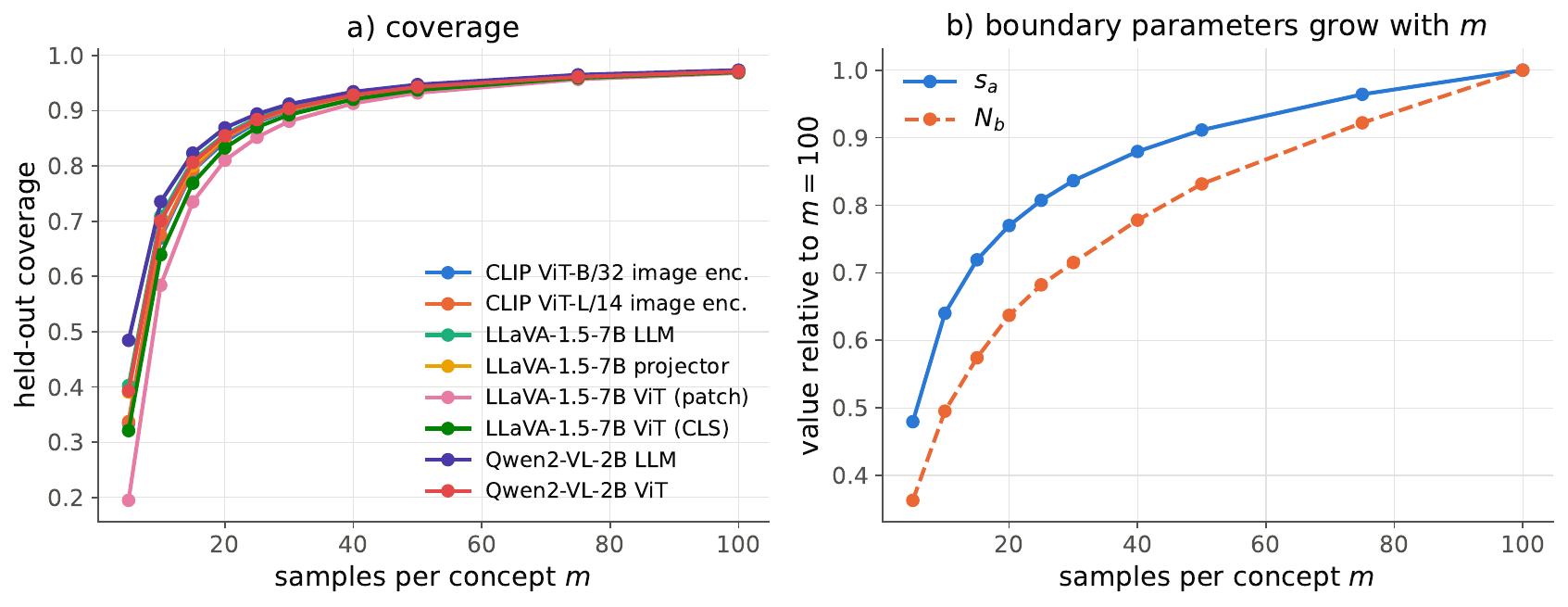}
    \vspace{-6mm}
    \caption{\textbf{Effect of the number of samples $m$ on fitted capsules.} Results are averaged over $796$ visual concepts, eight model components, three representative layers per component, and three independent draws. \textbf{a)} Held-out coverage increases with $m$ and gradually saturates as more samples are used for capsule matching. \textbf{b)} Axis span $s_a$ and norm band $N_b$, normalized by their values at $m=100$, increase with $m$ as the fitted capsule progressively covers the concept geometry.}
    \label{fig:sample-size}
\end{figure}

Held-out coverage increases rapidly with $m$, reaching $0.878$ at $m=25$, $0.941$ at $m=50$, and $0.971$ at $m=100$ (Figure~\ref{fig:sample-size}). The fitted axis span and norm band also increase with $m$, as larger samples progressively capture more of the concept's angular and norm extent. In particular, the $m=50$ fitting size used in our main experiments already achieves a mean held-out coverage of $0.941$, while increasing the sample size further provides diminishing gains.

We further evaluate the sampling stability of the fitted axis by independently matching two capsules to disjoint sets of $m$ samples from the same concept and measuring the cosine distance $\Delta a$ between their axes. The estimated axes become increasingly consistent as $m$ grows, with mean $\Delta a$ decreasing from $0.1150$ at $m=5$ to $0.0141$ at $m=50$ and $0.0071$ at $m=100$. Table~\ref{tab:sample-size-components} further shows that this trend is consistent across model components. These values provide a reference scale for interpreting the axis drifts measured across training checkpoints.

\begin{table}[h]
\centering
\caption{\textbf{Held-out coverage, axis span, and axis dispersion across model components.} Results are computed over $796$ visual concepts and three representative layers per component. Axis dispersion $\Delta a$ is the cosine distance between axes independently fitted from two disjoint sample sets of the same concept.}
\label{tab:sample-size-components}
\begin{tabular}{llrrrrrrr}
\toprule
& & \multicolumn{3}{c}{Coverage} & \multicolumn{2}{c}{$s_a$} & \multicolumn{2}{c}{$\Delta a$} \\
\cmidrule(lr){3-5}\cmidrule(lr){6-7}\cmidrule(lr){8-9}
Model & Component & $m{=}25$ & $m{=}50$ & $m{=}100$ & $m{=}25$ & $m{=}100$ & $m{=}25$ & $m{=}100$ \\
\midrule
CLIP ViT-B/32 & image enc.   & 0.879 & 0.942 & 0.972 & 0.146 & 0.175 & 0.0107 & 0.0027 \\
CLIP ViT-L/14 & image enc.   & 0.882 & 0.943 & 0.972 & 0.214 & 0.265 & 0.0142 & 0.0036 \\
Qwen2-VL-2B   & ViT          & 0.884 & 0.943 & 0.971 & 0.412 & 0.530 & 0.0277 & 0.0072 \\
Qwen2-VL-2B   & LLM          & 0.894 & 0.947 & 0.973 & 0.398 & 0.478 & 0.0176 & 0.0044 \\
LLaVA-1.5-7B  & ViT (patch)  & 0.852 & 0.932 & 0.968 & 0.646 & 0.745 & 0.0739 & 0.0198 \\
LLaVA-1.5-7B  & ViT (CLS)    & 0.870 & 0.937 & 0.969 & 0.273 & 0.323 & 0.0252 & 0.0065 \\
LLaVA-1.5-7B  & projector    & 0.884 & 0.943 & 0.973 & 0.507 & 0.640 & 0.0271 & 0.0068 \\
LLaVA-1.5-7B  & LLM          & 0.887 & 0.944 & 0.972 & 0.482 & 0.665 & 0.0231 & 0.0058 \\
\bottomrule
\end{tabular}
\end{table}

\subsection{Robustness to Capsule Boundary Estimation}
\label{sec:validation-boundary}

Capsule matching defines its boundary using the extrema of the fitting samples: $s_a=\max_i d_{\cos}(x_i,a)$, $n_{\mathrm{low}}=\min_i\|x_i\|$, and $n_{\mathrm{high}}=\max_i\|x_i\|$. To examine whether our results depend on this choice, we replace these extrema with percentile-based boundaries parameterized by $q$. We set $s_a$ to the $q$-th percentile of the angular distances and $(n_{\mathrm{low}},n_{\mathrm{high}})$ to the central $q\%$ interval of the sample norms, while keeping the fitted axis unchanged. We evaluate $q\in\{100,95,90,85,80\}$, where $q=100$ recovers the original capsule. We perform this analysis across different types of concepts and evaluate each fitted capsule using held-out coverage and disentanglement $\mathrm{Dis}$.

\begin{figure}[h]
    \centering
    \includegraphics[width=\linewidth]{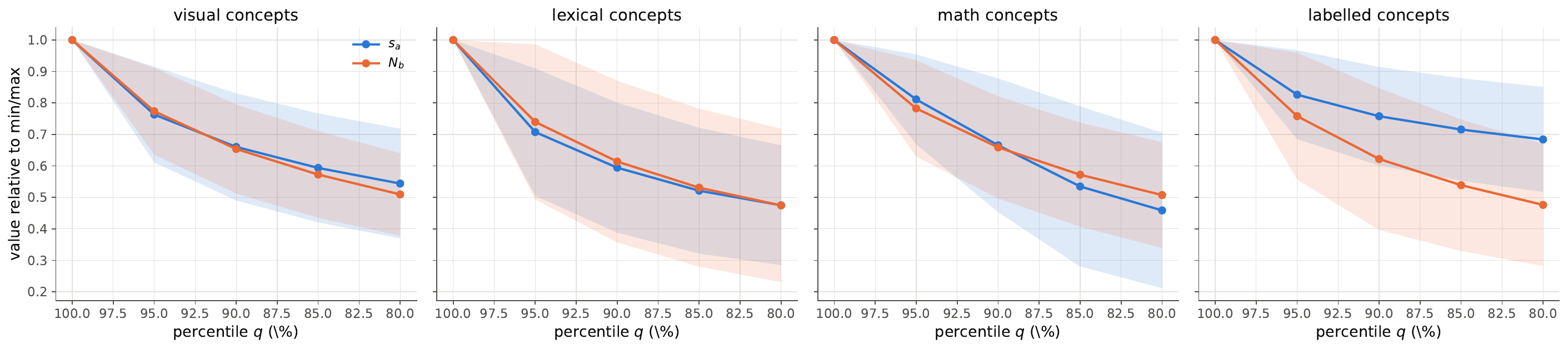}
    \vspace{-6mm}
    \caption{\textbf{Capsule parameters under percentile-based boundaries.} Axis span $s_a$ and norm band $N_b$ are normalized by their values at $q=100$ and averaged across concepts. Shaded regions indicate $\pm1$ standard deviation. Lower percentiles produce progressively tighter capsule boundaries across all four concept types.}
    \label{fig:boundary-parameters}
\end{figure}

As shown in Figure~\ref{fig:boundary-parameters}, both the axis span and norm band decrease consistently as the percentile is lowered. The largest contraction occurs between $q=100$ and $q=95$: at $q=95$, the mean axis span is $0.708$--$0.826$ of its original value across concept types, while the mean norm band is $0.740$--$0.783$ of its original value. Thus, the extreme samples have a substantial effect on the fitted boundary, but the same qualitative contraction is observed across modalities and concept constructions.

\begin{figure}[h]
    \centering
    \includegraphics[width=\linewidth]{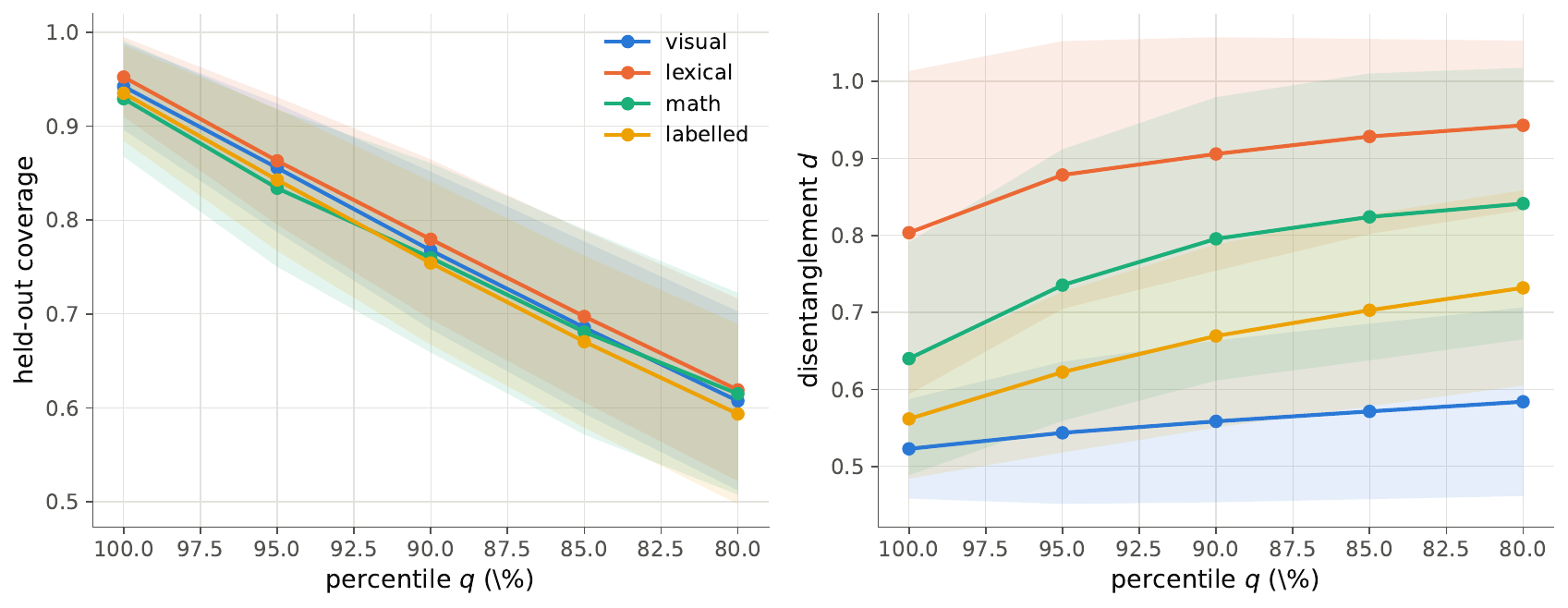}
    \vspace{-6mm}
    \caption{\textbf{Held-out coverage and disentanglement under percentile-based boundaries.} Held-out coverage (left) and disentanglement $\mathrm{Dis}$ (right) as the percentile $q$ decreases from the original min/max boundary ($q=100$) to increasingly tighter capsules. Results are averaged across concepts, with shaded regions indicating $\pm1$ standard deviation.}
    \label{fig:boundary-tradeoff}
\end{figure}

Tighter boundaries induce a consistent trade-off between coverage and disentanglement (Figure~\ref{fig:boundary-tradeoff}). Reducing $q$ from $100$ to $80$ decreases held-out coverage by approximately $0.32$--$0.34$ across all four concept types, while increasing disentanglement by $0.061$ for visual, $0.140$ for lexical, $0.201$ for math, and $0.170$ for labelled concepts. These results show that percentile-based boundaries can substantially increase the selectivity of fitted capsules, but only at the cost of excluding a considerable fraction of unseen positive samples. We therefore retain the min/max boundary in \CapsuleLens{}, prioritizing generalization to unseen samples while reporting disentanglement separately to characterize overlap with other concepts.

\subsection{Alternative Geometric Approximations under Different Boundaries}
\label{sec:validation-shapes}

We further examine whether the comparison with alternative geometric approximations in Section~\ref{sec:geometry-shapes} persists under percentile-based boundaries. A \emph{cone} retains the capsule axis $a$ and angular span $s_a$ without constraining the norm, while a \emph{band} retains only the norm interval $[n_{\mathrm{low}},n_{\mathrm{high}}]$. A \emph{ball} is centered at the mean of the fitting samples with radius $r=\max_i\|x_i-c\|$, and a \emph{box} uses the coordinate-wise minima and maxima of the fitting samples. All shapes are fitted to the same $50$ samples and evaluated using the same concepts, layers, held-out samples, and negatives.

We apply the percentile-based boundary construction of Section~\ref{sec:validation-boundary} to each geometric approximation. Table~\ref{tab:geometry-shapes-percentile} reports held-out coverage and disentanglement across $q\in\{100,95,90,85,80\}$. Tightening the boundaries produces the expected coverage--disentanglement trade-off for the capsule, cone, and ball. Across all five percentiles, the capsule consistently achieves higher disentanglement than the cone and ball, while the cone and ball retain higher coverage at the same $q$. The band remains substantially less disentangled across all boundary choices, with $\mathrm{Dis}$ increasing only from $0.514$ to $0.532$, further indicating that norm information alone provides limited separation. The box continues to exhibit extremely low held-out coverage across all settings. Overall, the results reveal a consistent pattern across boundary choices: directional information accounts for most of the observed concept separation, while incorporating the norm constraint yields a modest additional increase in disentanglement. The capsule therefore provides a compact parameterization that jointly captures both aspects while maintaining high held-out coverage under its original boundary.

\begin{table}[h]
\centering
\caption{\textbf{Alternative geometric approximations under percentile-based boundaries.} Held-out coverage (Cov.) and disentanglement $\mathrm{Dis}$, averaged across concept types, model components, and representative layers. Lower $q$ corresponds to a tighter boundary; $q=100$ recovers the original min/max construction.}
\label{tab:geometry-shapes-percentile}
\small
\setlength{\tabcolsep}{4pt}
\begin{tabular}{lcccccccccc}
\toprule
& \multicolumn{2}{c}{$q=100$} & \multicolumn{2}{c}{$q=95$} & \multicolumn{2}{c}{$q=90$} & \multicolumn{2}{c}{$q=85$} & \multicolumn{2}{c}{$q=80$} \\
\cmidrule(lr){2-3}\cmidrule(lr){4-5}\cmidrule(lr){6-7}\cmidrule(lr){8-9}\cmidrule(lr){10-11}
Shape & Cov. & $\mathrm{Dis}$ & Cov. & $\mathrm{Dis}$ & Cov. & $\mathrm{Dis}$ & Cov. & $\mathrm{Dis}$ & Cov. & $\mathrm{Dis}$ \\
\midrule
Capsule & 0.943 & \textbf{0.583} & 0.855 & \textbf{0.620} & 0.769 & \textbf{0.641} & 0.686 & \textbf{0.658} & 0.609 & \textbf{0.673} \\
Cone    & 0.975 & 0.578 & 0.918 & 0.614 & 0.862 & 0.635 & 0.805 & 0.651 & 0.750 & 0.665 \\
Ball    & 0.977 & 0.568 & 0.922 & 0.601 & 0.865 & 0.620 & 0.811 & 0.635 & 0.757 & 0.649 \\
Band    & 0.962 & 0.514 & 0.916 & 0.521 & 0.867 & 0.526 & 0.818 & 0.529 & 0.769 & 0.532 \\
Box     & 0.012 & 0.653 & 0.003 & 0.456 & 0.003 & 0.890 & 0.002 & 1.000 & 0.002 & 1.000 \\
\bottomrule
\end{tabular}
\end{table}
\newpage
\section{Case Study I: Tracking Concept Geometry in CLIP Pretraining}
\label{app:clip-pretraining}

This appendix extends the main-text discussion of Section~\ref{sec:drift-pretraining}
and Figure~\ref{fig:clip-pretraining}, giving additional views on the
same training run and checkpoints — the within-concept distribution behind
the block-11 span contraction, and the concept-level analysis of the
drift signal.

\subsection{Axis-span contraction is population-wide}
\label{subsec:app-exp1-span-norm}

\begin{figure}[htbp]
  \centering
  \includegraphics[width=0.7\linewidth]{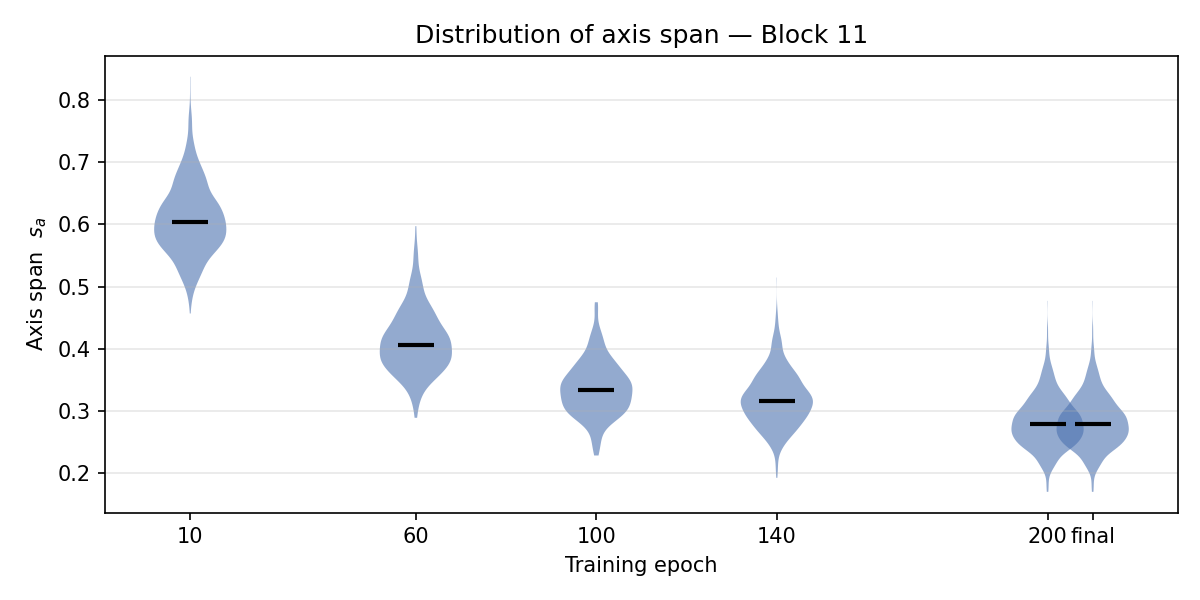}
  \caption{\textbf{Block-11 span contraction is visible across the concept
  distribution.} Distribution of axis-span $s_a$ across concepts at block
  11, by training epoch. The distribution shifts downward (mean
  $\approx 0.60 \to 0.28$) and narrows over training, without a separate
  non-contracting subpopulation appearing at any checkpoint.}
  \label{fig:clip-sa-violin}
\end{figure}

Figure~\ref{fig:clip-sa-violin} shows that the block-11 span contraction
reported in the main text holds across the concept population, rather than
being driven by a subset of concepts.

\subsection{Concept-level structure}
\label{subsec:app-exp1-concepts}

\begin{figure}[htbp]
  \centering
  \includegraphics[width=0.8\linewidth]{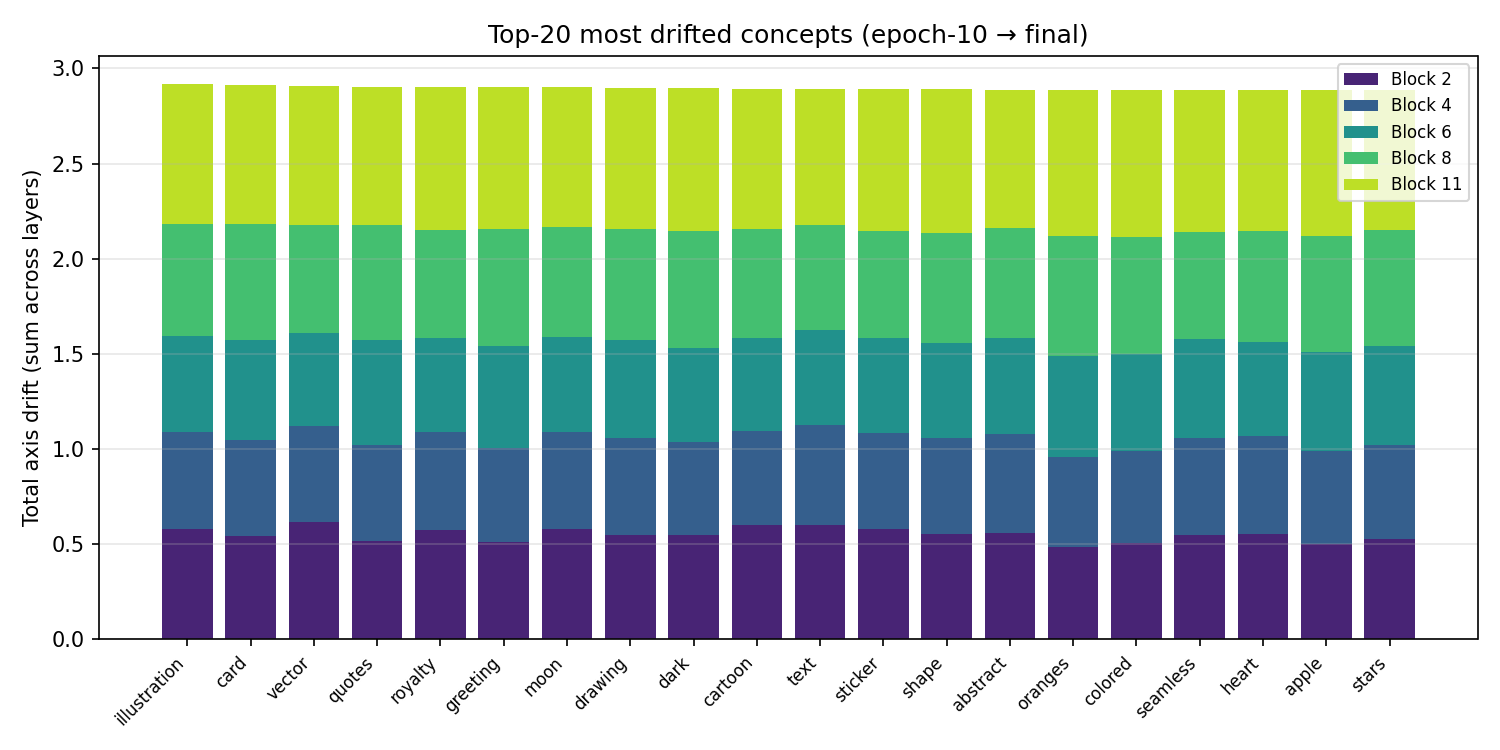}
  \caption{\textbf{Top-20 most drifted concepts, block 11.} Total axis
  drift (epoch-10 $\to$ final) summed across the five probed blocks
  (stacked bars, one color per block), for the 20 concepts with the
  largest totals. Totals span a narrow range ($\approx 2.88$--$2.92$).}
  \label{fig:clip-top-drifted}
\end{figure}

\begin{figure}[htbp]
  \centering
  \includegraphics[width=0.8\linewidth]{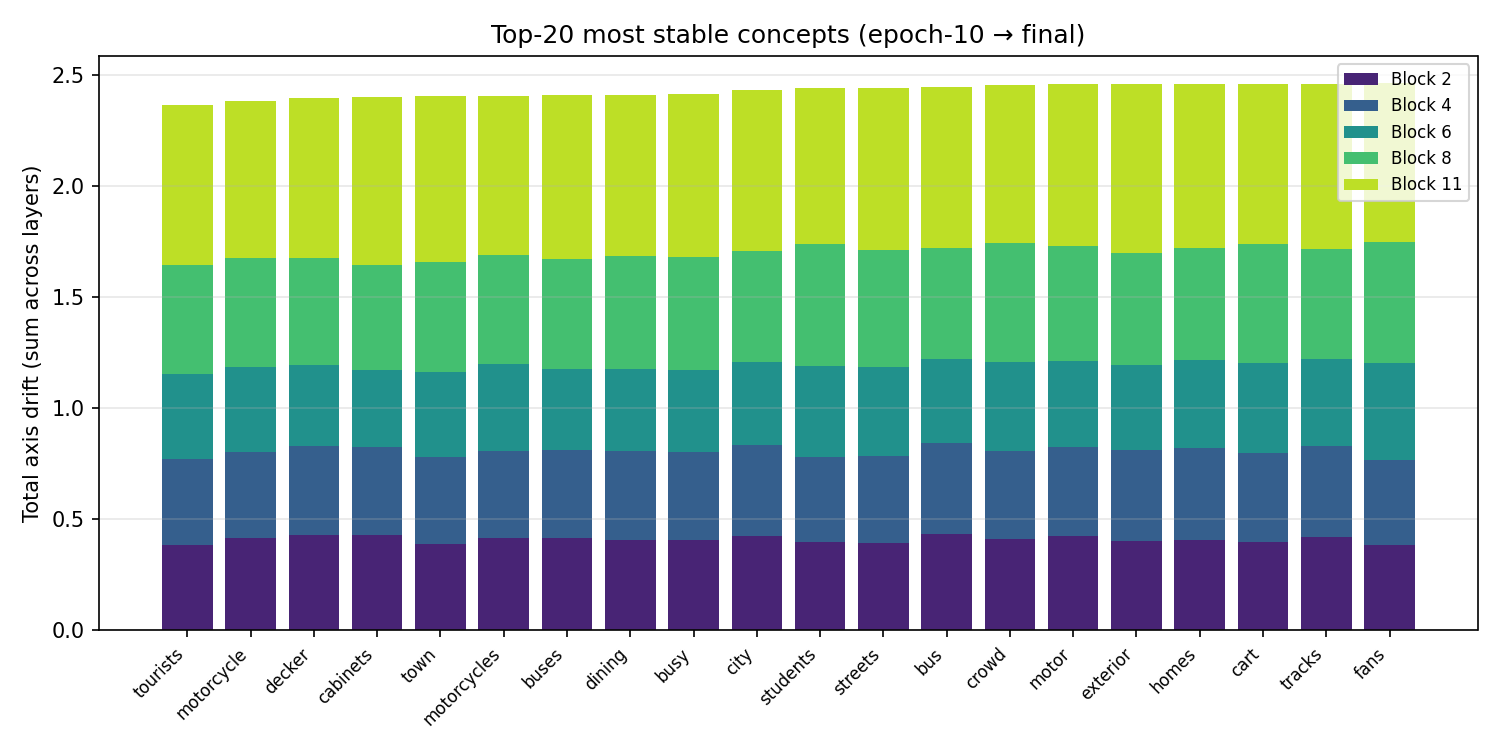}
  \caption{\textbf{Top-20 most stable concepts, block 11.} Total axis
  drift (epoch-10 $\to$ final) summed across the five probed blocks, for
  the 20 concepts with the smallest totals. Totals ($\approx 2.36$--$2.46$)
  are only $\approx 20\%$ below the most-drifted group in
  Figure~\ref{fig:clip-top-drifted}, indicating that the gap between the
  two extremes is present but modest relative to the overall drift scale.}
  \label{fig:clip-least-drifted}
\end{figure}

Figures~\ref{fig:clip-top-drifted}--\ref{fig:clip-least-drifted} break down
the block-11 rotation reported in the main text by concept, illustrating
that Capsule Lens can rank individual concepts by drift magnitude within a
single probed block.

\subsection{Axis drift vs.\ span change}
\label{subsec:app-exp1-span}

\begin{figure}[htbp]
  \centering
  \includegraphics[width=0.6\linewidth]{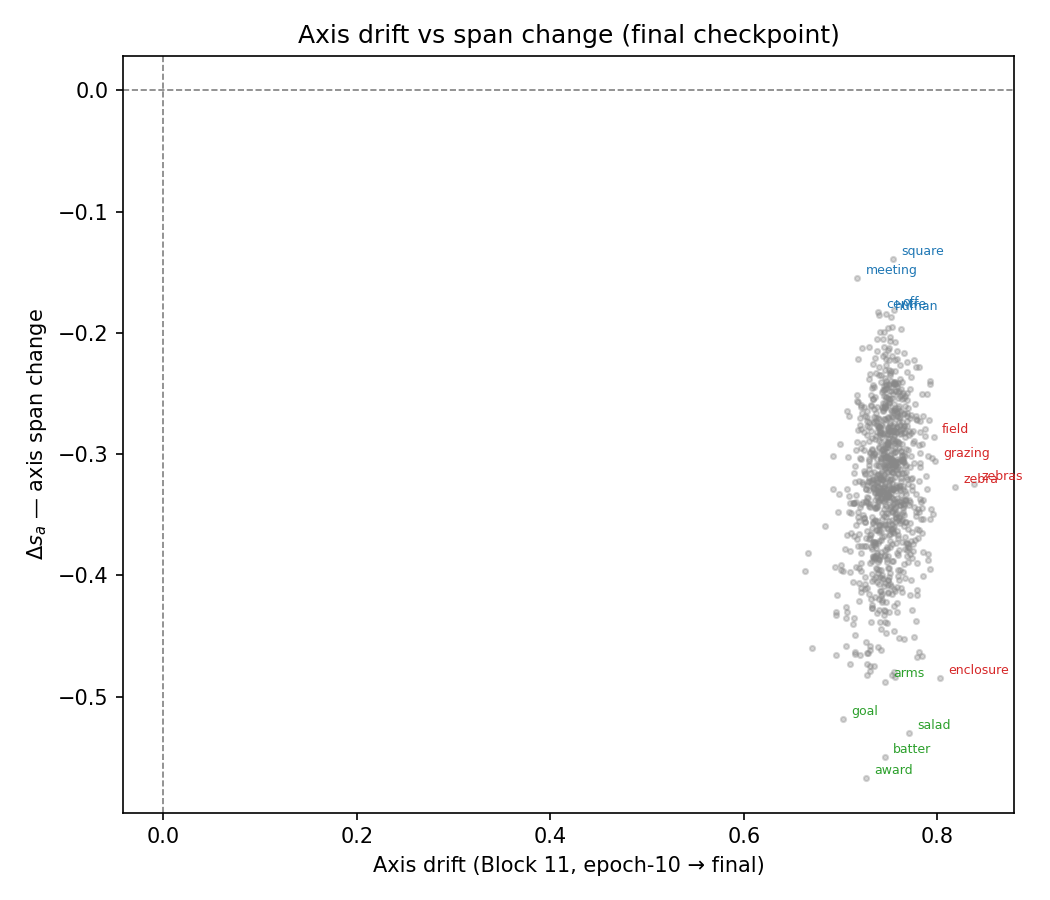}
  \caption{\textbf{Axis drift and span change at block 11, final
  checkpoint.} Axis drift vs.\ span change $\Delta s_a$ for individual
  concepts, with five example concepts labeled at each of three regions of
  the plot: lowest span change in magnitude (blue: \emph{square},
  \emph{meeting}, \emph{coffee}, \emph{human}), highest axis drift (red:
  \emph{field}, \emph{grazing}, \emph{zebra}, \emph{zebras},
  \emph{enclosure}), and largest span change in magnitude (green:
  \emph{arms}, \emph{goal}, \emph{salad}, \emph{batter}, \emph{award}).
  All plotted concepts fall in a narrow drift band ($\approx 0.65$--$0.87$)
  and have $\Delta s_a < 0$ ($\Delta s_a \in [-0.57, -0.14]$).}
  \label{fig:clip-drift-sa-quadrants}
\end{figure}

Figure~\ref{fig:clip-drift-sa-quadrants} relates the block-11 span
contraction to the per-concept drift of Section~\ref{subsec:app-exp1-concepts},
jointly plotting both quantities for individual concepts.

\newpage
\section{Case Study II: PPO for VLM on Visual Question Answering}
\label{sec:app-exp2}

This appendix extends the main-text discussion of PPO post-training on
Qwen2-VL, giving additional views on the same training run and
checkpoints — the concept-level breakdown of the layer-27 drift signal,
and the joint relationship between axis drift and span change.

\subsection{Concept-level heterogeneity}
\label{subsec:app-exp2-concepts}

\begin{figure}[htbp]
  \centering
  \includegraphics[width=0.8\linewidth]{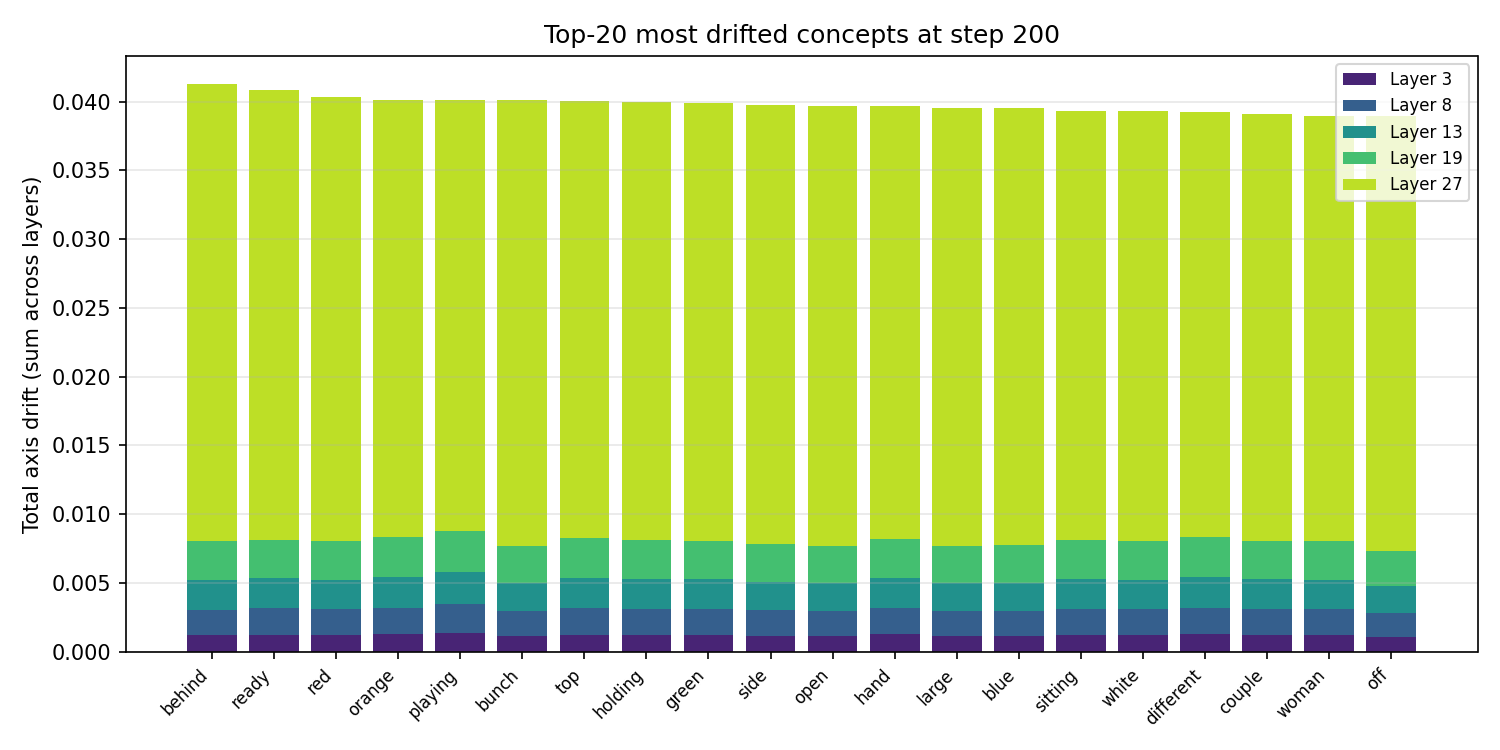}
  \caption{\textbf{Top-20 most drifted concepts, layer 27.} Total axis
  drift at step 200, summed across the five probed layers (stacked bars,
  one color per layer), for the 20 concepts with the largest totals. Layer
  27 (top segment) accounts for most of each total.}
  \label{fig:top-drifted}
\end{figure}

\begin{figure}[htbp]
  \centering
  \includegraphics[width=0.8\linewidth]{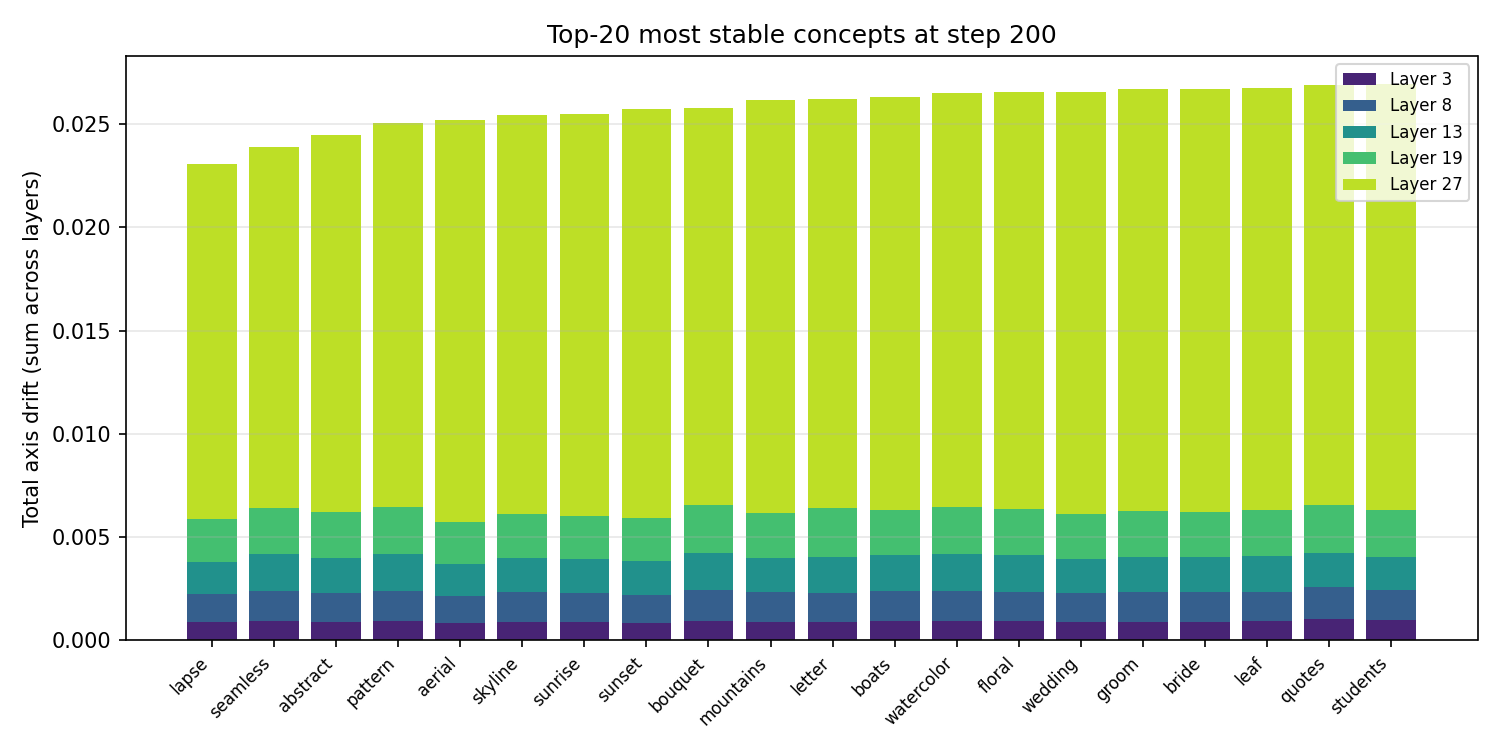}
  \caption{\textbf{Top-20 most stable concepts, layer 27.} Total axis
  drift at step 200, summed across the five probed layers, for the 20
  concepts with the smallest totals. The y-axis maximum ($\approx 0.027$)
  is close to Figure~\ref{fig:top-drifted}'s ($\approx 0.041$), indicating
  the gap between most- and least-drifted concepts is present but modest.}
  \label{fig:least-drifted}
\end{figure}

Figures~\ref{fig:top-drifted}--\ref{fig:least-drifted} rank individual
concepts by total axis drift at layer 27, the layer where drift
concentrates.

\begin{figure}[htbp]
  \centering
  \includegraphics[width=0.8\linewidth]{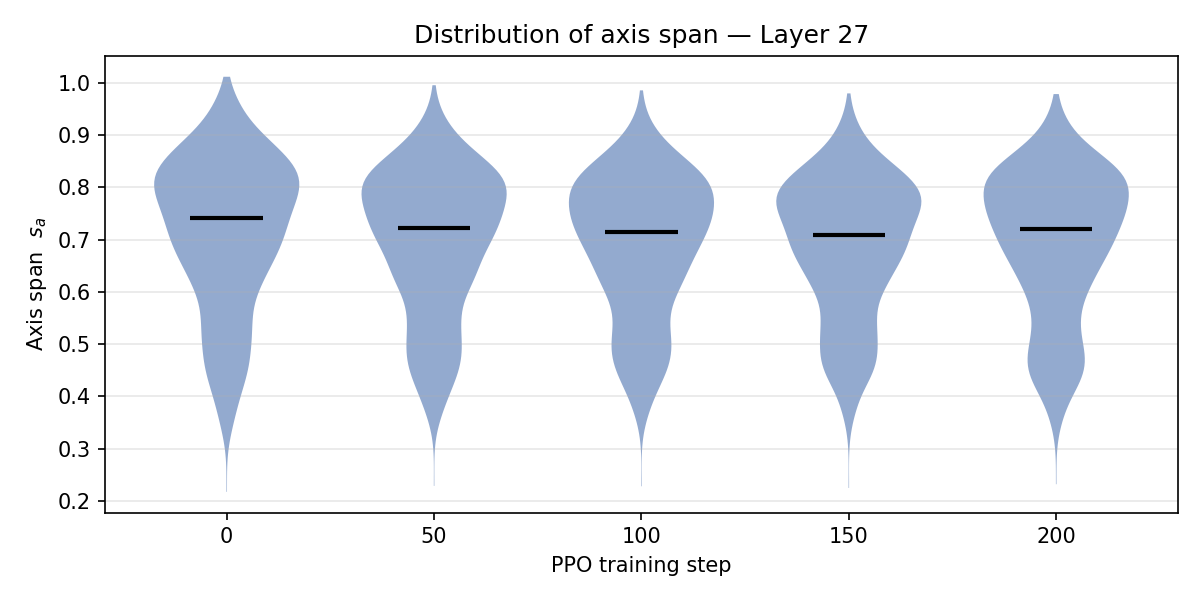}
  \caption{\textbf{Axis-span distribution at layer 27 across checkpoints.}
  Distribution of axis-span $s_a$ across concepts at layer 27, by PPO
  training step. The distribution shape is similar across all five
  checkpoints; the mean varies between $\approx 0.71$ and $\approx 0.74$.}
  \label{fig:sa-violin}
\end{figure}

\subsection{Axis drift vs.\ span change}
\label{subsec:app-exp2-span}

\begin{figure}[htbp]
  \centering
  \includegraphics[width=0.6\linewidth]{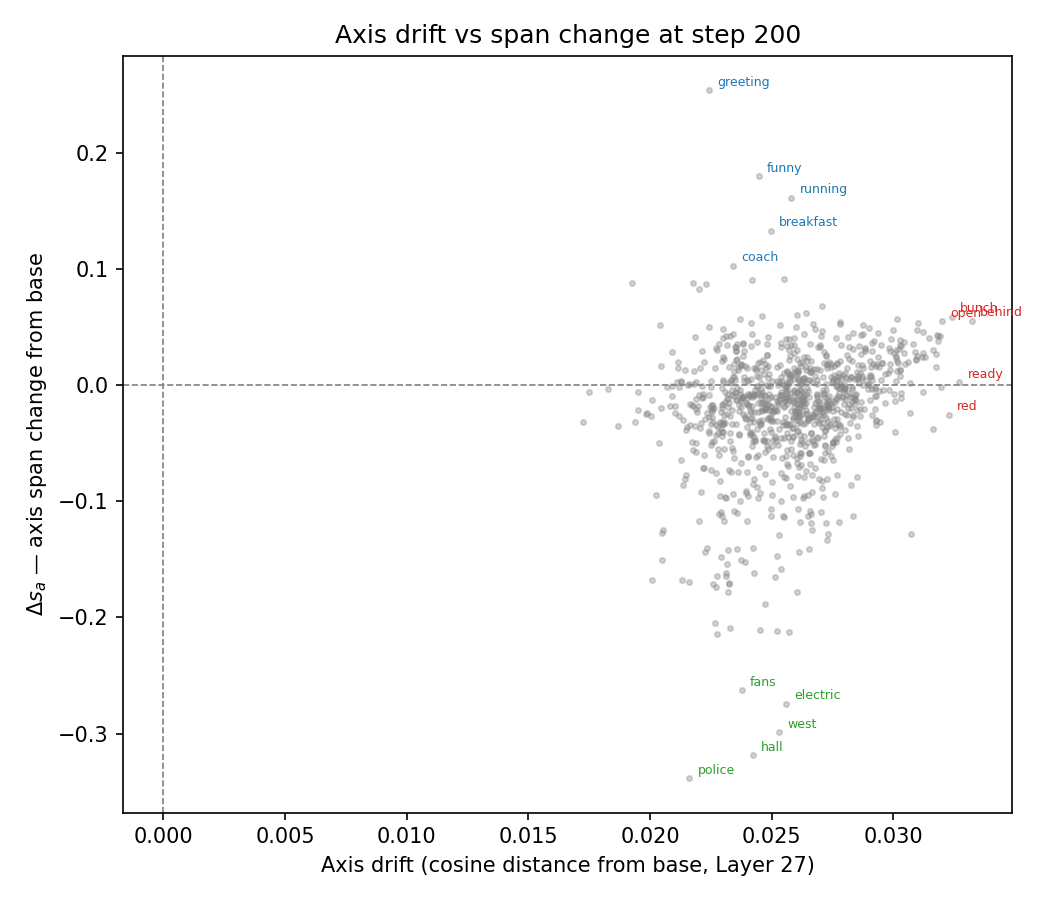}
  \caption{\textbf{Axis drift and span change at layer 27, final
  checkpoint.} Axis drift vs.\ span change $\Delta s_a$ (step 200 minus
  base) for individual concepts, with five example concepts labeled at
  each of three regions of the plot: largest positive span change (blue:
  \emph{greeting}, \emph{funny}, \emph{running}, \emph{breakfast},
  \emph{coach}), lowest span change in magnitude (red: \emph{bunch},
  \emph{open}, \emph{behind}, \emph{ready}, \emph{red}), and largest
  negative span change (green: \emph{fans}, \emph{electric}, \emph{west},
  \emph{hall}, \emph{police}). Axis drift for all plotted concepts falls
  between $0.018$ and $0.033$.}
  \label{fig:drift-sa-quadrants}
\end{figure}

Figure~\ref{fig:drift-sa-quadrants} jointly plots axis drift and span
change $\Delta s_a$ for individual concepts at layer 27, the final
checkpoint.

\begin{figure}[H]
  \centering
  \includegraphics[width=0.8\linewidth]{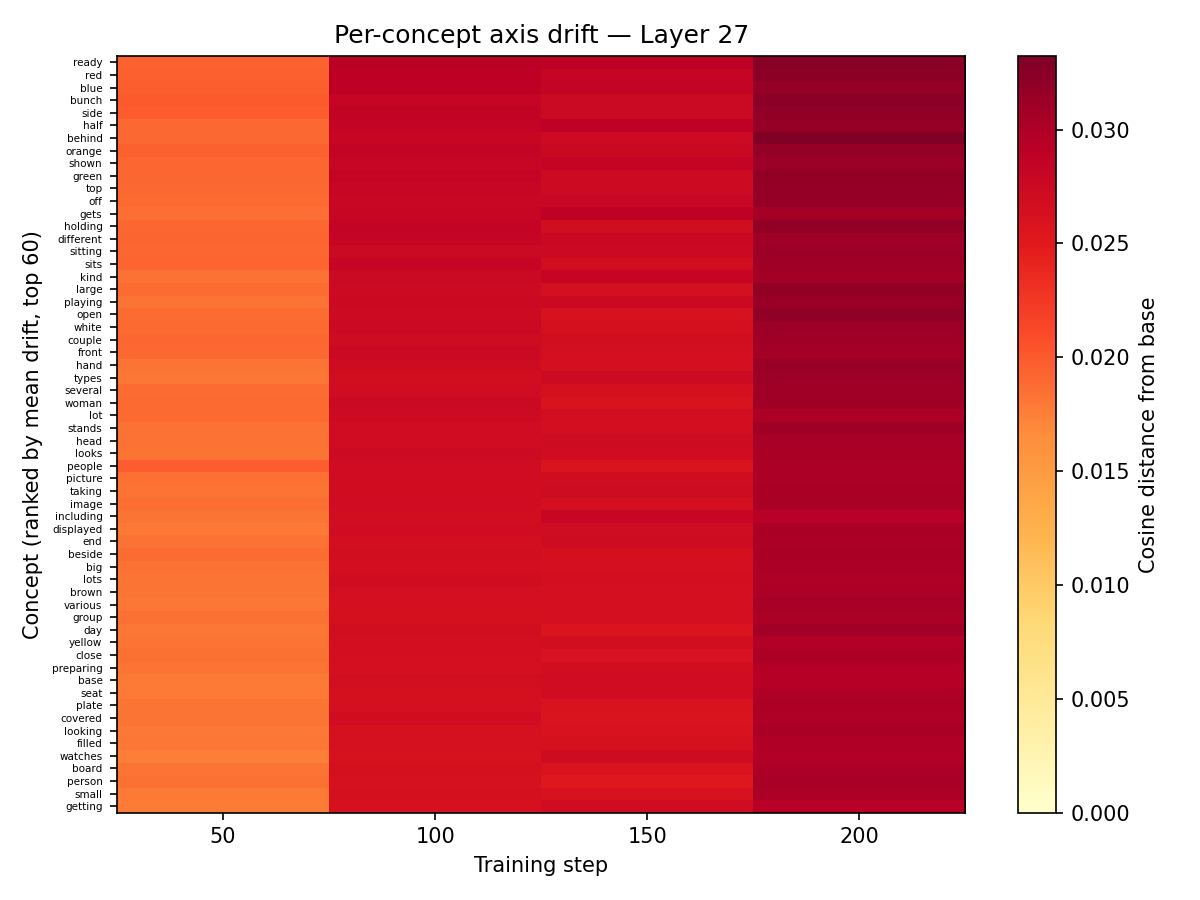}
  \caption{\textbf{Per-concept axis drift at layer 27 across training.}
  Rows are the top 60 concepts by mean drift, columns are PPO training
  steps. Color encodes cosine distance from the base axis.}
  \label{fig:drift-heatmap}
\end{figure}

\newpage
\section{Case Study III: RLVR on Mathematical Reasoning}
\label{app:case3}

This appendix provides additional analyses for the RLVR experiment in Section~\ref{sec:drift-case3}. We examine both the population-level evolution of concept geometry and the individual trajectories of all $18$ mathematical concepts under the same GRPO training run. Capsules are fitted using $50$ samples per concept and evaluated on a disjoint set of $50$ samples; held-out coverage averages $0.942$ across training, ranging from $0.80$ to $1.00$.

\subsection{Population-Averaged Geometry}
\label{app:case3-averaged}

\begin{figure}[H]
\centering
\includegraphics[width=\textwidth]{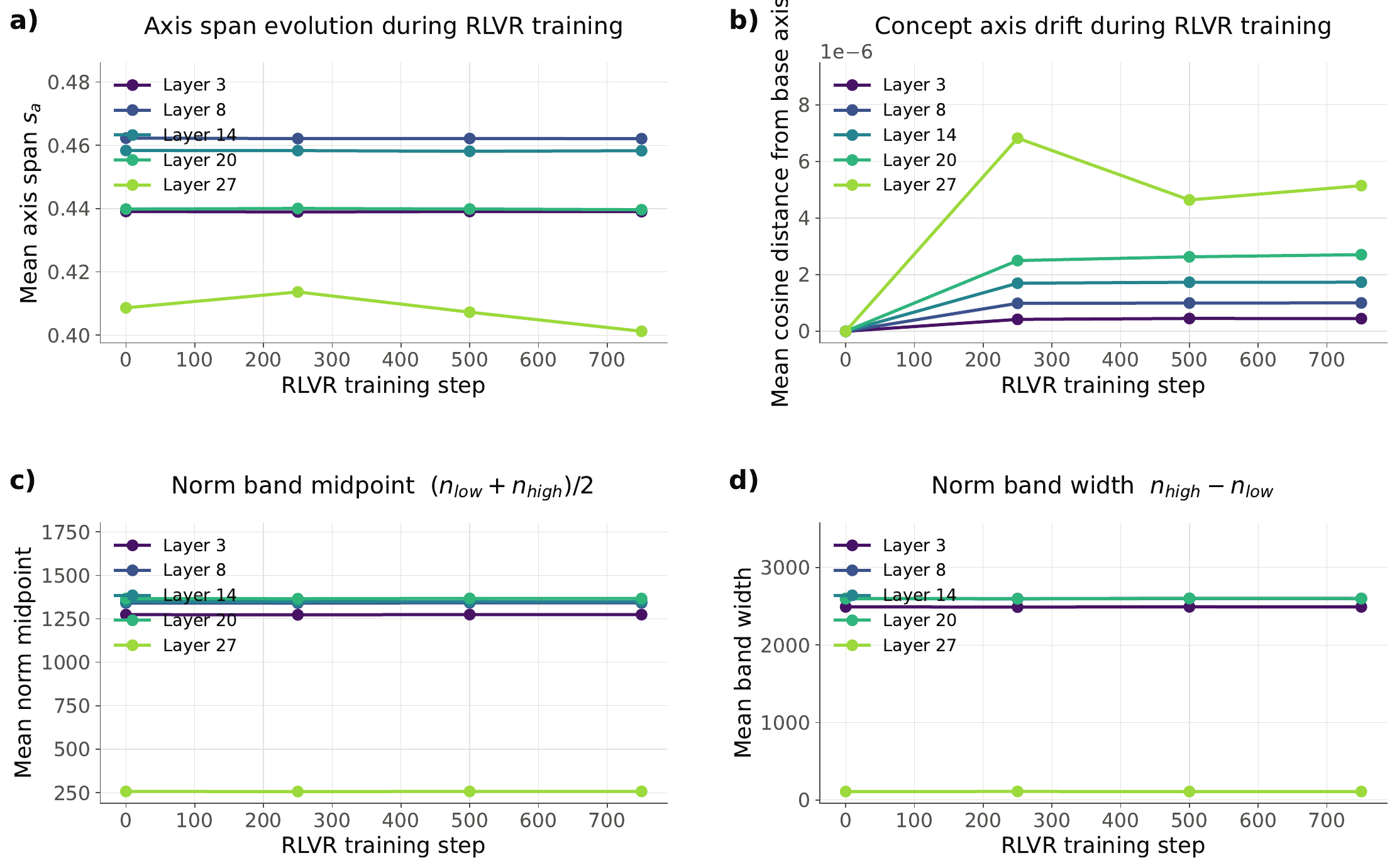}
\caption{\textbf{Population-averaged concept geometry during RLVR.} \textbf{(a)} Mean axis span across the $18$ mathematical concepts remains nearly constant at layers $3$--$20$, while layer $27$ shows a small late contraction. \textbf{(b)} Mean axis drift from the base model remains below $7\times10^{-6}$ at every layer. \textbf{(c,d)} Mean norm-band midpoint and width remain largely unchanged throughout training.}
\label{fig:rlvr-averaged}
\end{figure}

We first average each capsule parameter across the $18$ mathematical concepts at every probed layer. Figure~\ref{fig:rlvr-averaged} shows that population-level geometry remains highly stable under RLVR. Mean axis drift stays below $7\times10^{-6}$ at every layer, and axis spans remain nearly unchanged at layers $3$--$20$; only layer $27$ exhibits a small non-monotonic change followed by a modest contraction. Mean norm-band midpoint and width are similarly stable throughout training. These averages therefore indicate that RLVR induces little global restructuring of concept geometry.

The population average can nevertheless conceal changes concentrated in a small number of concepts. We therefore inspect the trajectory of each mathematical concept separately.

\subsection{Per-Concept Trajectories}
\label{app:case3-perconcept}

Per-concept trajectories reveal a strongly heterogeneous response to RLVR. At layer $27$, only $7$ of the $18$ concepts change their axis span by more than $0.005$, while \texttt{unit\_time\_minute} exhibits by far the largest contraction and is analyzed in Figure~\ref{fig:rlvr-specificity} of the main text. The following figures report axis drift, axis span, norm-band midpoint, and norm-band width for the remaining concepts at all five probed layers.

\begin{figure}[H]
\centering
\includegraphics[width=0.80\linewidth]{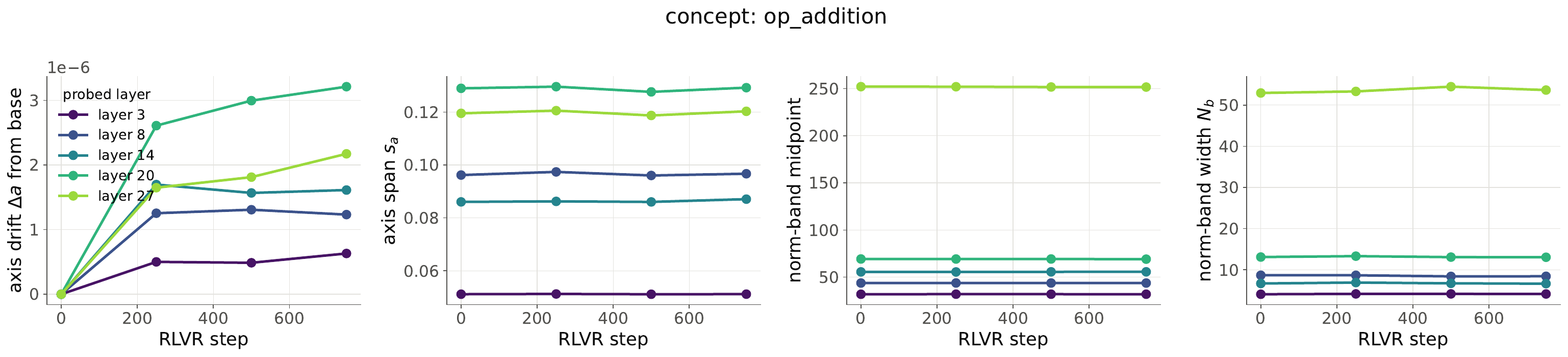}\\[-1mm]
\includegraphics[width=0.80\linewidth]{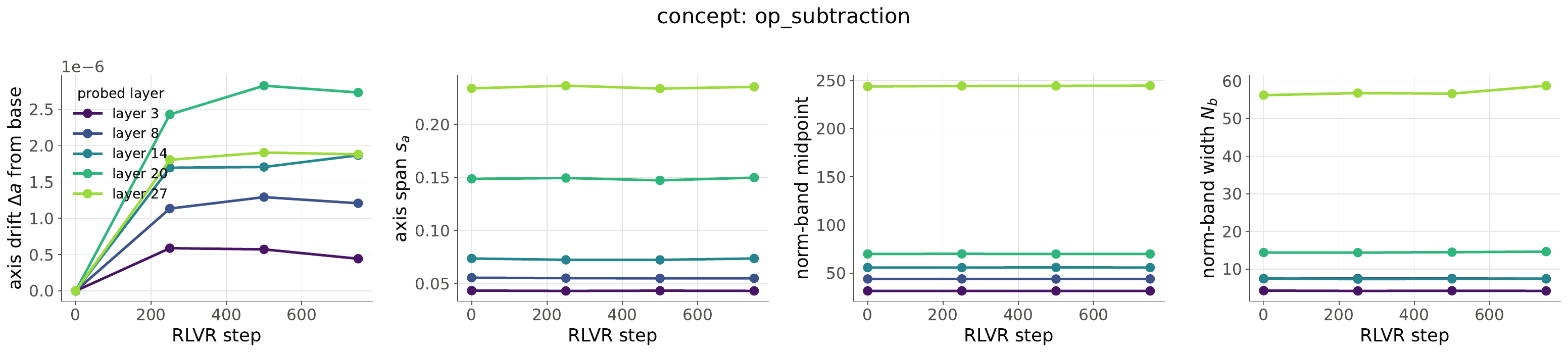}\\[-1mm]
\includegraphics[width=0.80\linewidth]{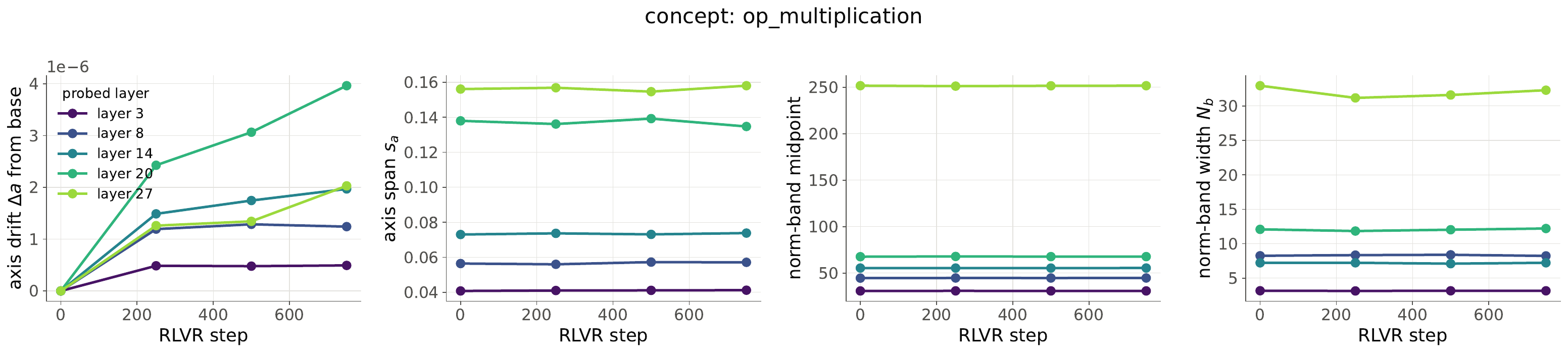}\\[-1mm]
\includegraphics[width=0.80\linewidth]{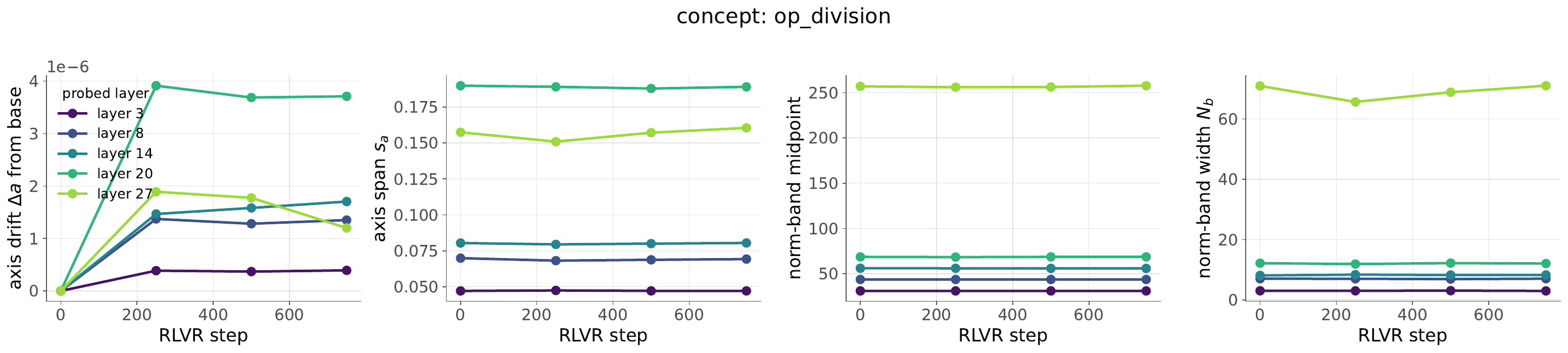}
\caption{\textbf{RLVR trajectories of arithmetic-operation concepts (I).} The concepts \texttt{op\_addition}, \texttt{op\_subtraction}, \texttt{op\_multiplication}, and \texttt{op\_division} remain largely stable across training, with only minor changes in axis span and norm geometry.}
\label{fig:rlvr-ops-1}
\end{figure}

\begin{figure}[H]
\centering
\includegraphics[width=0.80\linewidth]{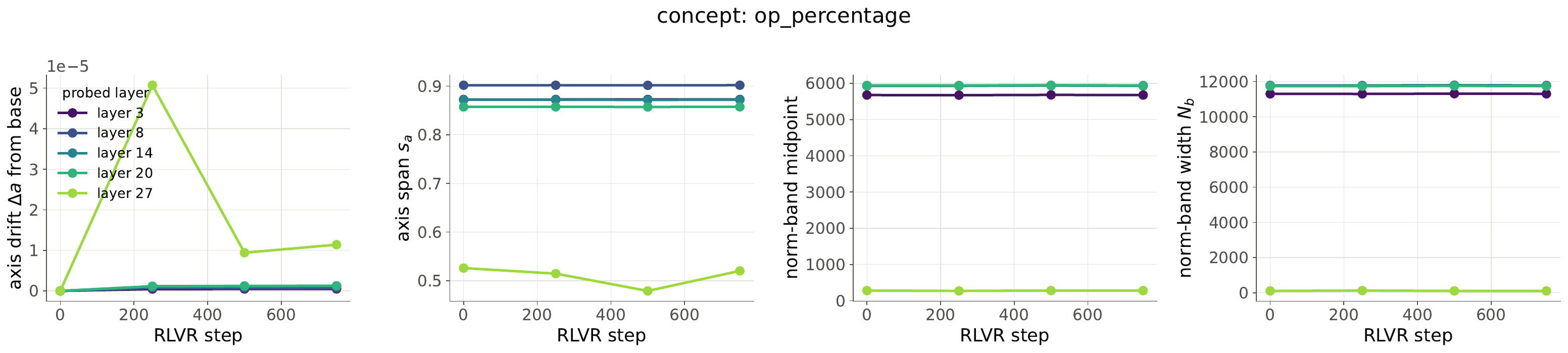}\\[-1mm]
\includegraphics[width=0.80\linewidth]{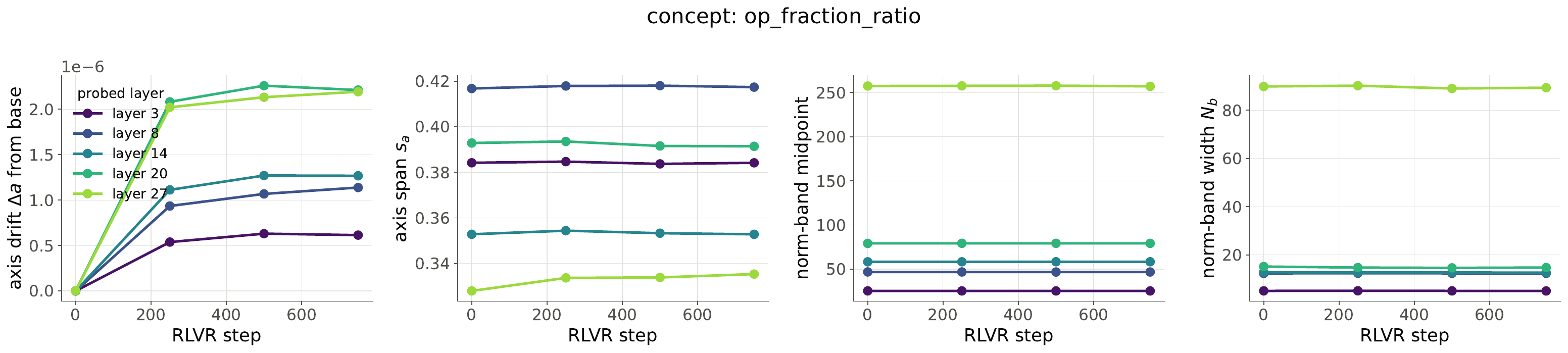}\\[-1mm]
\includegraphics[width=0.80\linewidth]{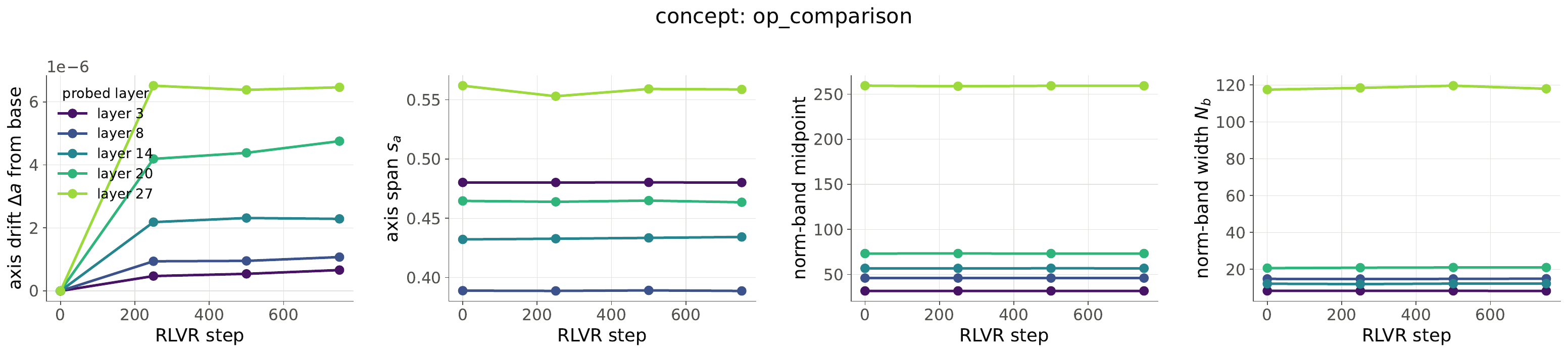}
\caption{\textbf{RLVR trajectories of arithmetic-operation concepts (II).} The concepts \texttt{op\_percentage}, \texttt{op\_fraction\_ratio}, and \texttt{op\_comparison} also exhibit limited geometric change, although \texttt{op\_percentage} shows somewhat larger variation than most other operation concepts. Overall, arithmetic-operation concepts remain geometrically stable under RLVR.}
\label{fig:rlvr-ops-2}
\end{figure}

Taken together, these trajectories explain why the population averages remain nearly unchanged despite several visible outliers: RLVR does not broadly restructure the mathematical concept space, but selectively modifies the geometry of a small number of concepts at particular layers. This provides additional evidence for the concept- and layer-specific concentration effect highlighted in Section~\ref{sec:drift-case3}.

\begin{figure}[H]
\centering
\includegraphics[width=0.80\linewidth]{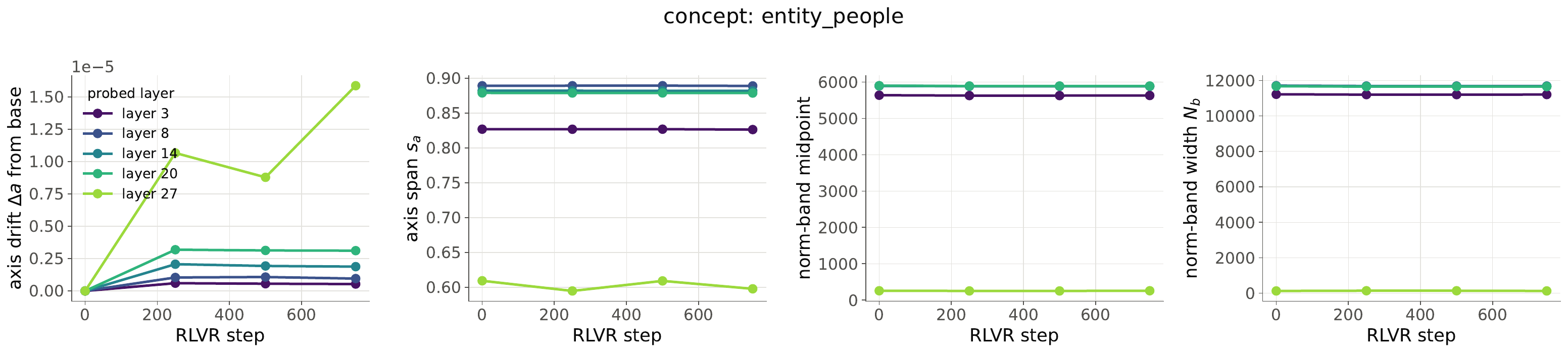}\\[-1mm]
\includegraphics[width=0.80\linewidth]{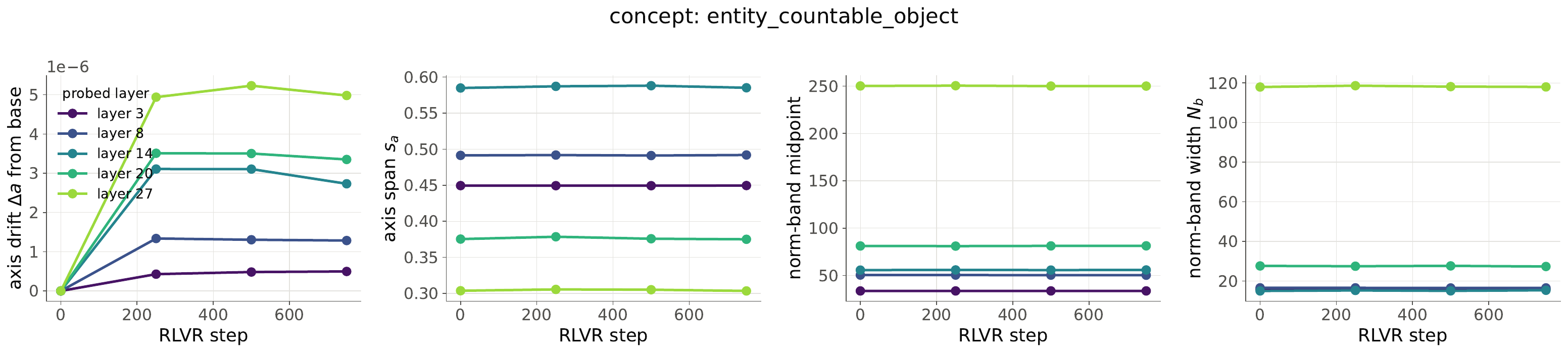}
\caption{\textbf{RLVR trajectories of entity concepts.} \texttt{entity\_people} exhibits one of the clearest geometric changes outside \texttt{unit\_time\_minute}, including a layer-$27$ span contraction and substantial norm-bound changes at intermediate layers. In contrast, \texttt{entity\_countable\_object} remains comparatively stable, further illustrating the concept-specific nature of RLVR-induced geometric change.}
\label{fig:rlvr-entities}
\end{figure}

\begin{figure}[H]
\centering
\includegraphics[width=0.80\linewidth]{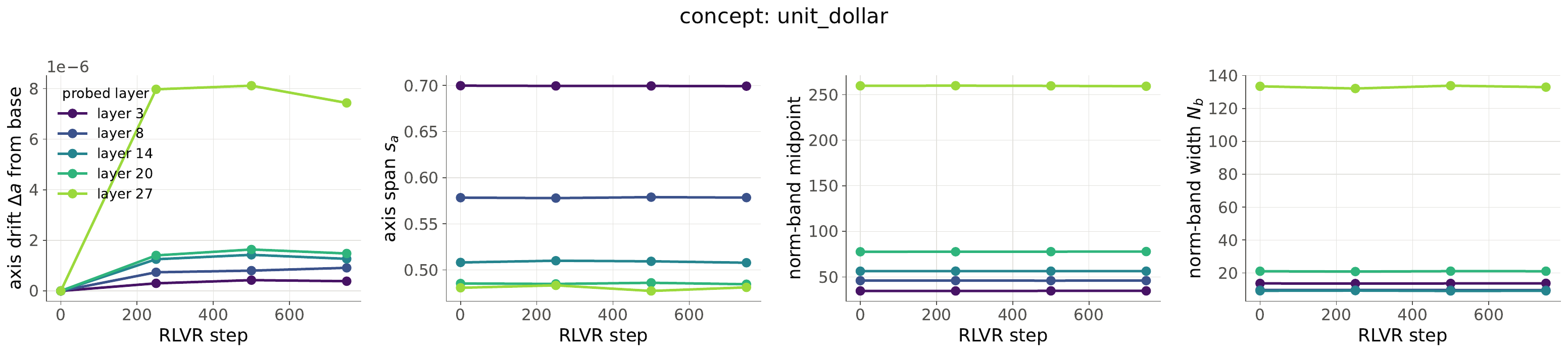}\\[-1mm]
\includegraphics[width=0.80\linewidth]{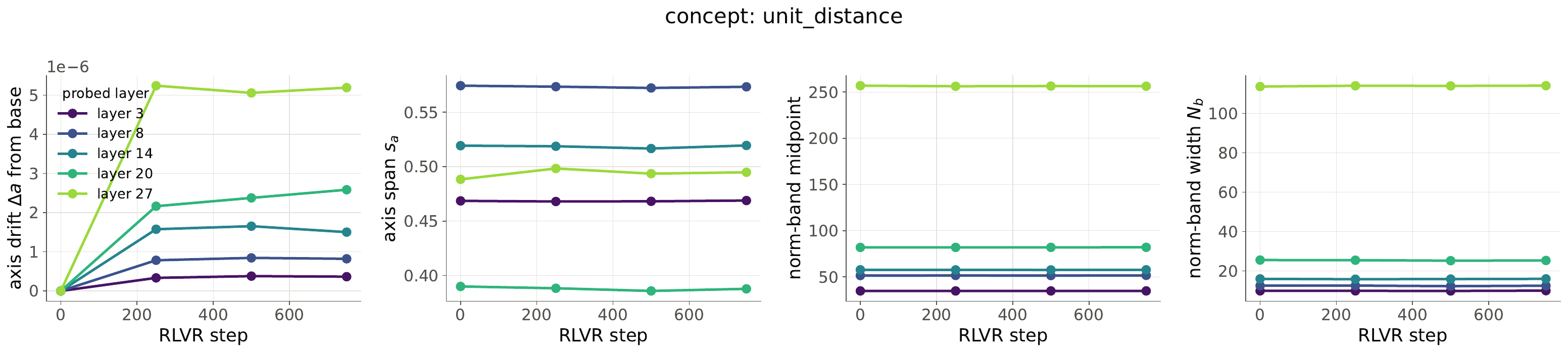}\\[-1mm]
\includegraphics[width=0.80\linewidth]{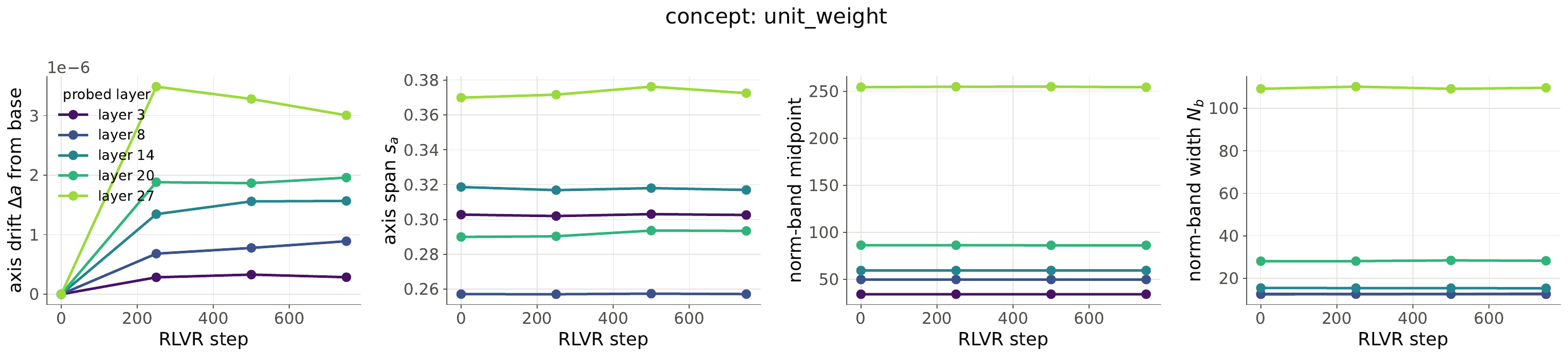}\\[-1mm]
\includegraphics[width=0.80\linewidth]{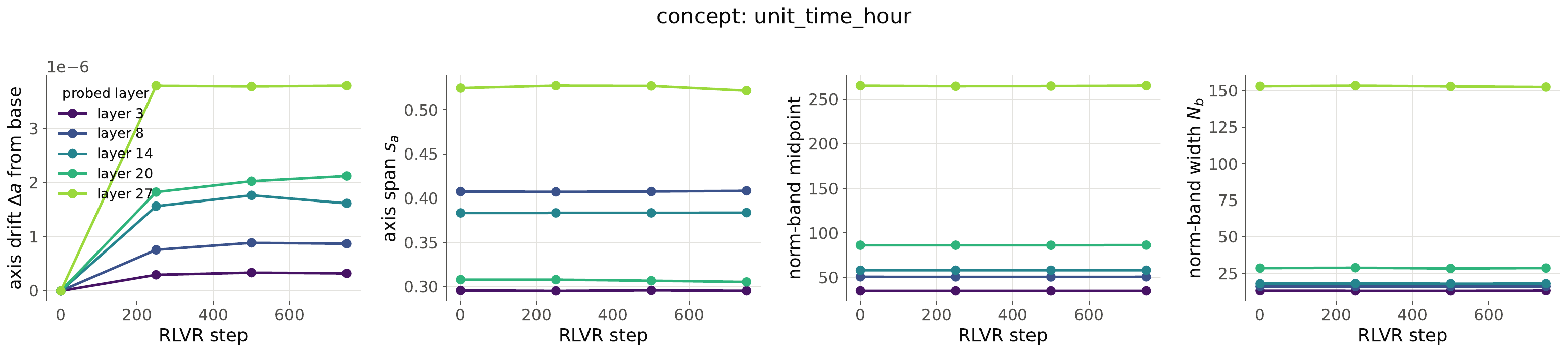}
\caption{\textbf{RLVR trajectories of unit concepts (I).} The concepts \texttt{unit\_dollar}, \texttt{unit\_distance}, \texttt{unit\_weight}, and \texttt{unit\_time\_hour} remain largely stable throughout training, with only small changes in axis span and norm geometry.}
\label{fig:rlvr-units-1}
\end{figure}

\begin{figure}[H]
\centering
\includegraphics[width=0.80\linewidth]{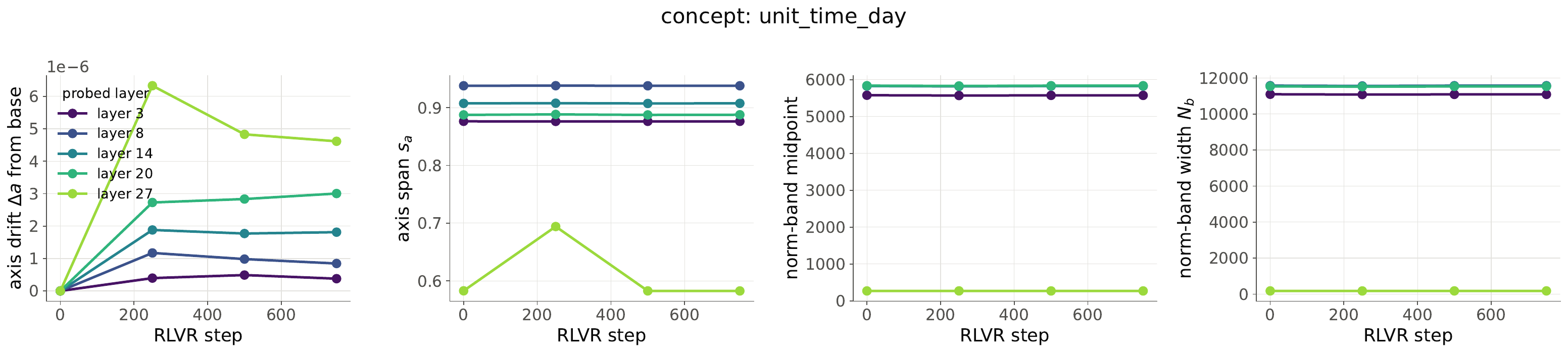}\\[-1mm]
\includegraphics[width=0.80\linewidth]{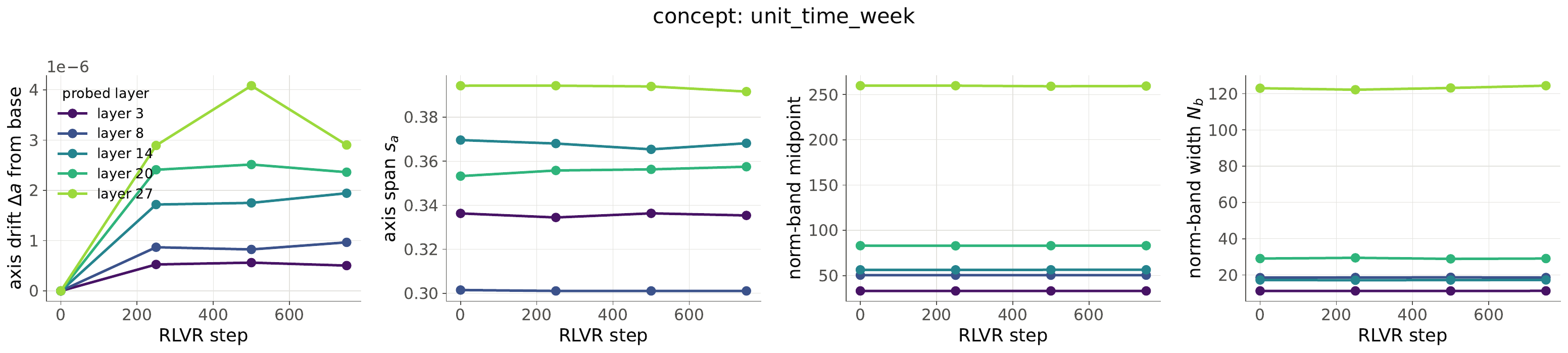}\\[-1mm]
\includegraphics[width=0.80\linewidth]{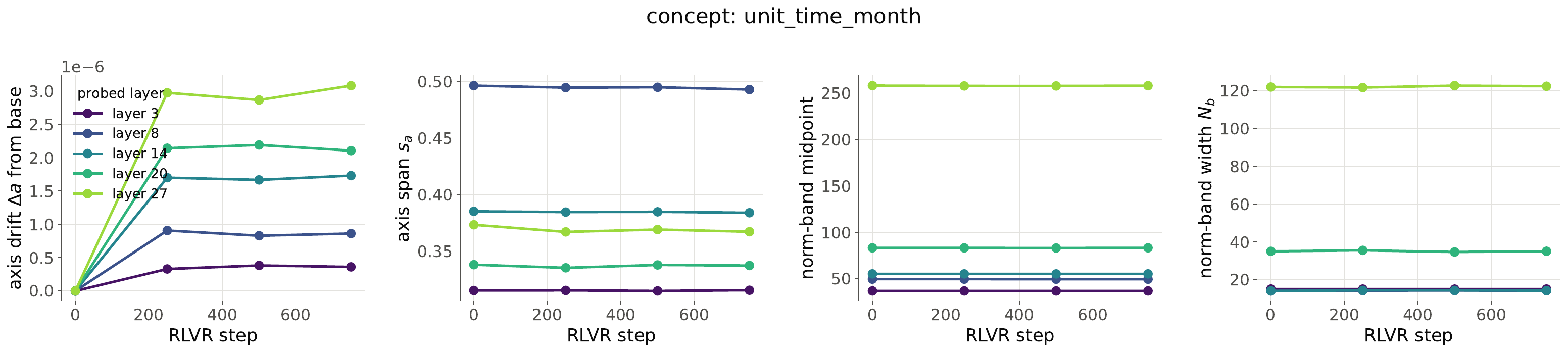}\\[-1mm]
\includegraphics[width=0.80\linewidth]{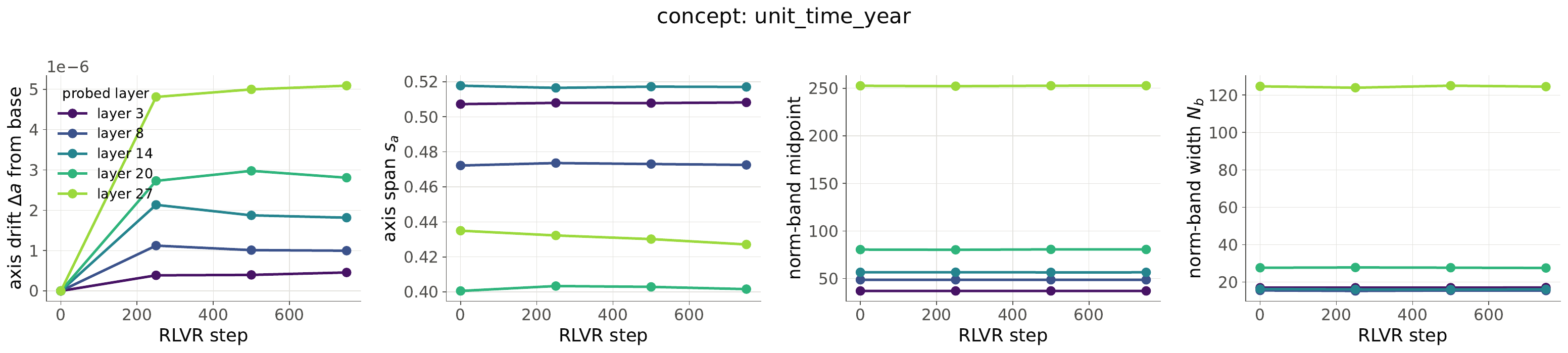}
\caption{\textbf{RLVR trajectories of unit concepts (II).} The concepts \texttt{unit\_time\_day}, \texttt{unit\_time\_week}, \texttt{unit\_time\_month}, and \texttt{unit\_time\_year} exhibit heterogeneous but generally limited changes. Among them, \texttt{unit\_time\_month} and \texttt{unit\_time\_year} show the largest remaining layer-$27$ span contractions, but both are substantially smaller than the contraction observed for \texttt{unit\_time\_minute} in the main text. This further indicates that RLVR-induced concentration is concept-specific rather than shared uniformly across the time-unit family.}
\label{fig:rlvr-units-2}
\end{figure}
\newpage
\section{Geometry-Guided Representation Steering with Capsule Lens}
\label{app:steering}

Beyond locating and tracking concept geometry, we test whether the geometry identified by \CapsuleLens{} can also provide useful directions for representation steering. For a concept whose span curve exhibits multimodal structure, we apply spherical $k$-means to its unit-normalized representations to obtain mode centroids. Given source and target centroids $\mu_i$ and $\mu_j$, we define the unit steering direction $d_{i\rightarrow j}=(\mu_j-\mu_i)/\|\mu_j-\mu_i\|$. We use the fitted norm shell to calibrate the intervention magnitude, setting $m=(\nlow+\nhigh)/2$ and $\lambda=km$ with $k\in[0.6,0.9]$, and intervene as $h'=h+\lambda d_{i\rightarrow j}$. When the two centroids correspond to different modes within one concept, we call this \emph{intra-concept steering}; when they correspond to different concepts, we call it \emph{inter-concept steering}.

\paragraph{Intra-concept steering.} Table~\ref{tab:intra-steering} shows representative examples for polysemous textual concepts. Steering toward opposite modes can selectively elicit different meanings of the same concept, suggesting that the modes identified within a capsule correspond to behaviorally meaningful semantic structure.

\begin{table}[h]
\centering
\caption{\textbf{Intra-concept steering between semantic modes.} Representative generations from Qwen2.5-1.5B after steering toward different modes within the same concept geometry.}
\label{tab:intra-steering}
\small
\begin{tabular}{lll}
\toprule
\textbf{Concept} & \textbf{Target mode} & \textbf{Steered generation} \\
\midrule
church & institution & ``a community of all the members of Christ... authority in the Church'' \\
church & building & ``a big, old building... red door with a clock on it'' \\
plant & factory & ``a facility for processing... waste gas emissions for steel production'' \\
plant & organism & ``long, bushy stems... grow up to 10 inches tall in their garden'' \\
state & nation & ``located in the United States... population'' \\
state & condition & ``the state of a system... an evolving and dynamic one'' \\
cell & programming & ``a variable in the language of programming'' \\
\bottomrule
\end{tabular}
\end{table}

\paragraph{Inter-concept steering.} We further construct directions between centroids belonging to different concepts. Table~\ref{tab:inter-steering} shows that the same geometric construction can shift generations toward distinct textual and visual concepts.

\begin{table}[h]
\centering
\caption{\textbf{Inter-concept steering across textual and visual representations.} Each example is shown with its base generation followed by the generation obtained after steering toward the target concept.}
\label{tab:inter-steering}
\small
\begin{tabular}{lll}
\toprule
\textbf{Modality} & \textbf{Direction} & \textbf{Generation} \\
\midrule
Text & church $\rightarrow$ home & \textbf{Base:} ``a place of worship... pray, sing hymns'' \\
     &                         & \textbf{Steered:} ``a place where people can relax and enjoy comfort...'' \\
\midrule
Text & cell $\rightarrow$ plant & \textbf{Base:} ``a small unit of life... DNA and proteins'' \\
     &                          & \textbf{Steered:} ``it has its own life... doesn't need anything else to grow'' \\
\midrule
Vision & dog $\rightarrow$ cat & \textbf{Base:} ``two dogs resting on a cushion in a car'' \\
       &                       & \textbf{Steered:} ``two cats resting on a fluffy surface...'' \\
\midrule
Vision & dog $\rightarrow$ cat & \textbf{Base:} ``two dogs in the back seat of a car...'' \\
       &                       & \textbf{Steered:} ``two cats, one orange and one black-and-white...'' \\
\midrule
Vision & animal $\rightarrow$ vehicle & \textbf{Base:} ``two dogs in the back seat of a car...'' \\
       &                              & \textbf{Steered:} ``two vehicles parked on the side of the road...'' \\
\midrule
Vision & animal $\rightarrow$ vehicle & \textbf{Base:} ``two dogs on a cushion in a car'' \\
       &                              & \textbf{Steered:} ``a parked vehicle... a large black van with a trailer'' \\
\bottomrule
\end{tabular}
\end{table}

Together, these qualitative results suggest that the geometry recovered by \CapsuleLens{} can provide actionable directions for representation intervention: internal modes support steering between different meanings of the same concept, while relative concept locations support steering across concepts. We leave systematic evaluation of this capability to future work.

\end{document}